\documentclass[11pt]{article}
\usepackage{acl}
\usepackage[T1]{fontenc}
\usepackage[utf8]{inputenc}
\usepackage{times}
\usepackage{latexsym}
\usepackage{microtype}
\usepackage{amsmath,amssymb,amsthm,mathtools}
\usepackage{booktabs}
\usepackage{array}
\usepackage{graphicx}
\usepackage{xcolor}
\usepackage{algorithm}
\usepackage{algorithmic}
\usepackage{hyperref}
\usepackage{caption}
\usepackage{hyperref}
\usepackage{xurl}
\hypersetup{colorlinks=true,linkcolor=blue,citecolor=blue,urlcolor=blue}

\newtheorem{theorem}{Theorem}
\newtheorem{proposition}[theorem]{Proposition}
\newtheorem{lemma}[theorem]{Lemma}
\newtheorem{corollary}[theorem]{Corollary}
\theoremstyle{definition}

\newtheorem{assumption}[theorem]{Assumption}
\theoremstyle{remark}

\newcommand{\R}{\mathbb{R}}
\newcommand{\E}{\mathbb{E}}
\newcommand{\Var}{\mathrm{Var}}
\newcommand{\Cov}{\mathrm{Cov}}
\newcommand{\Prb}{\mathbb{P}}
\newcommand{\calX}{\mathcal{X}}
\newcommand{\calA}{\mathcal{A}}

\newcommand{\norm}[1]{\left\lVert #1\right\rVert}
\newcommand{\one}{\mathbf{1}}

\newcolumntype{L}[1]{>{\raggedright\arraybackslash}p{#1}}

\title{Evaluator Ensembles Under Reward Hacking:\\
Covariance Geometry and Finite-Search Guarantees}

\author{
\textbf{Fariya Afrin}\textsuperscript{1}%
\thanks{Equal contribution.}%
\and
\textbf{Ibne Farabi Shihab}\textsuperscript{2}\footnotemark[1]
\thanks{Corresponding author: \texttt{ishihab@iastate.edu}.}
\\[2pt]
\textsuperscript{1}Department of Computer Science, Kalinga Institute of Industrial Technology \\
\textsuperscript{2}Department of Computer Science, Iowa State University \\
\texttt{ishihab@iastate.edu}
}

\date{}

\begin{document}
\maketitle

\begin{abstract}
Language-model judges and reward models enable scalable supervision, but finite optimization can exploit evaluator errors rather than improve response quality. We characterize this failure through the covariance geometry of evaluator ensembles. For calibrated judges, the ensemble mean retains common-mode error along the all-ones direction, whereas cross-judge disagreement captures only orthogonal error. Consequently, disagreement can be high despite robust aggregation, or low while shared response-dependent errors persist. We prove that common-mode error is not identifiable from internal judge scores alone. Under a joint sub-Gaussian model, we bound best-of-$K$ selection overstatement and target-quality regret, extending the guarantees to predictably adaptive search under conditional calibration. The resulting search terms scale as $\sqrt{\log K}$ and are asymptotically tight for Gaussian projected errors. We further show that noisy quality proxies introduce artificial rank-one covariance without changing disagreement, and propose a bounded two-anchor Bernstein certificate for finite-search error and regret. Fixed-seed Gaussian stress tests over 120 $(J,\rho,K)$ configurations and real-model audits validate the theory while revealing the limits of disagreement-based diagnostics under increasing search pressure.
\end{abstract}

\section{Introduction}

Learned reward models and language-model judges have become central to reinforcement learning from human feedback, direct preference optimization, reranking, and automatic evaluation \citep{christiano2017deep,stiennon2020learning,ouyang2022training,rafailov2023direct}. Their scalability comes with a structural vulnerability. Once a generator is optimized against a fixed evaluator, it can increase the evaluator's score by exploiting imperfections in the proxy rather than by improving the quality that the proxy was intended to represent. This behavior is commonly described as reward hacking, evaluator gaming, or reward overoptimization \citep{amodei2016concrete,gao2023scaling,skalse2022defining}.

An ensemble is a natural defense. If different judges make different mistakes, averaging their scores can prevent a response from succeeding by exploiting only one member. Empirical work confirms that ensembles often mitigate overoptimization, especially when their members differ before reward-model fine-tuning \citep{coste2024ensembles,eisenstein2023helping}. Yet the same studies also expose a limit: judges trained from related data, architectures, and preference signals can share the errors that optimization discovers. Adding more correlated judges then creates additional votes without adding commensurate information \citep{kim2025correlated,goel2025great,kohli2026nine}.

This paper makes the connection between error dependence and optimization pressure explicit. The key object is not the raw number of judges but the projection of their joint error onto the aggregation direction. For a uniform ensemble, this direction is the all-ones vector. The error component parallel to it changes the ensemble mean and can therefore be exploited by search. The orthogonal component makes judges disagree but cancels under averaging. This geometric view recovers the familiar equal-variance formula
\begin{equation}
\Var(\bar e)
=\sigma^2\left(\frac{1}{J}+\frac{J-1}{J}\rho\right),
\label{eq:intro-variance}
\end{equation}
but it also shows why that identity alone is not a reward-hacking result: an operational guarantee must concern the candidate selected by the proxy and the target-quality loss induced by that selection.

We therefore analyze a generator that draws $K$ candidates and selects the one with the highest ensemble score. Under a joint sub-Gaussian error model, we bound the selected response's proxy overstatement and its regret relative to the candidate with the highest target quality. The bound separates prompt-level offsets, which change absolute scores but not within-prompt selection, from response-dependent common-mode errors, which remain exploitable after averaging. It also makes the role of the search budget explicit. The guarantee is an upper bound of order $\sqrt{\log K}$ in general. For independent Gaussian projected errors, classical extreme-value theory makes this order tight for $\mathbb E\max_{k\leq K}\bar e_k$; it does not by itself establish tightness for selected-response overstatement or target-quality regret under arbitrary target qualities. We do not infer an exact empirical scaling law from the upper bound alone. A conditional version covers finite adaptive proposal sequences, making explicit the calibration obligation created when search reacts to earlier evaluator outputs.

The same geometry clarifies what disagreement measures. Disagreement is often used for active learning and uncertainty sampling \citep{seung1992query,houlsby2011bayesian,gal2017deep}, including for language reward models \citep{gleave2022uncertainty}. It is informative about the orthogonal, judge-specific component of error, but it is blind to an error shared by all judges. This limitation is compounded by a measurement problem: true response quality is rarely observed, so empirical studies often subtract one stronger judge as a proxy. We prove that independent noise in this quality proxy appears as a rank-one common mode in every estimated judge error. A two-anchor cross-covariance construction removes this particular contamination and separates the theoretical object from an artifact of the audit instrument.

Our contribution is consequently narrower than introducing ensembles, correlated-error analysis, or disagreement sampling. Gleave and Irving report that ensemble-based active learning for language reward models does not outperform random selection \citep{gleave2022uncertainty}; Eisenstein et al.\ show that shared reward-model errors survive ensembling \citep{eisenstein2023helping}; Kohli quantifies effective independence in discrete judge panels \citep{kohli2026nine}; CARE models latent confounders for static judge aggregation \citep{zhao2026care}; and concurrent work shows that pairwise correlation cannot identify the all-model co-failure tail in discrete orchestration \citep{chen2026combining}. Against this background, we provide an optimization-aware analysis for continuous evaluator scores. Specifically, we characterize the decomposition of ensemble error into common-mode and disagreement components and prove that the former is not identifiable from internal judge scores alone; derive finite and predictably adaptive search guarantees for selected-response overstatement and target-quality regret; and show how a noisy quality anchor induces a spurious rank-one common mode together with a finite-sample two-anchor correction. We evaluate these results using controlled Gaussian stress tests, a best-of-$K$ real-model audit, a multi-family finite-search audit over five eligible ensemble models from three families with a separately held-out Llama-family proxy excluded from every ensemble, two generators (\S\ref{sec:multifamily}), an anchor-sensitivity and verifier-backed evaluation (\S\ref{sec:targetquality}), and an exploratory DPO pilot reported in Appendix~\ref{sec:selection-result}. These experiments measure the quantities predicted by the theory and assess its practical implications, without claiming universality beyond the evaluated settings.

\section{Related Work}
\label{sec:related-work}

Reward overoptimization has been measured under best-of-$n$, policy optimization, and direct alignment algorithms \citep{gao2023scaling,rafailov2024scaling}.Complementary work detects proxy gaming through invariance-based evaluator
stress tests that separate sensitivity to exploitable features from
content-driven improvements under controlled perturbations
\citep{akter-etal-2026-detecting}. Whereas that work diagnoses whether
proxy-score gains arise from exploitable sensitivities, we characterize
which components of correlated evaluator error survive ensemble aggregation
and how best-of-$K$ search amplifies them. Ensemble defenses include mean aggregation, worst-case aggregation, uncertainty penalties, and weight averaging \citep{coste2024ensembles,rame2024warm}. Eisenstein et al. show that diversity introduced before reward-model fine-tuning is more useful than diversity introduced only during fine-tuning, while common qualitative failures remain exploitable by every ensemble member \citep{eisenstein2023helping}. Our search analysis is complementary: it does not propose that the mean is the optimal aggregator, but characterizes how error in a chosen aggregation direction is amplified by a finite candidate budget. The directly adjacent decision-level audit of \citet{landesberg2026bestofn} shows on a $5{,}000$-prompt Chatbot Arena best-of-$4$ benchmark that global judge--label correlation badly overstates within-prompt selection value, because global agreement is dominated by prompt-level baseline effects; that work diagnoses the score-versus-decision gap for a \emph{single} judge, whereas our contribution is the covariance geometry of a judge \emph{ensemble} under search---which error component averaging removes, what disagreement can and cannot certify, and a bounded variance certificate for the selected candidate---so the two are complementary rather than overlapping. Likewise, inference-time analyses of best-of-$n$ reward hacking and its guarantees \citep{gao2023scaling,huang2025bestofn} treat the proxy--true-reward gap for a given scalar proxy; we locate that gap in the common-mode component of an ensemble's error covariance and make it auditable.

Disagreement is a long-standing acquisition signal in query-by-committee and Bayesian active learning \citep{seung1992query,houlsby2011bayesian,gal2017deep}. In the closest reward-model study, \citet{gleave2022uncertainty} find that ensemble active learning does not beat random sampling and that estimated epistemic uncertainty is only weakly related to error. More recent work documents correlated failures across large model families \citep{kim2025correlated,goel2025great}, measures the effective number of votes in judge panels \citep{kohli2026nine}, and learns confounder-aware static aggregators \citep{zhao2026care}. Concurrent work on routing and voting further shows that pairwise error correlation does not identify the probability that all models fail on the same item \citep{chen2026combining}. That result concerns a discrete co-failure tail; our covariance is likewise not presented as a tail certificate. The distinction here is between static aggregation accuracy and optimization-induced selection under an explicit joint tail condition: even a small residual error can matter when a generator searches specifically for high proxy scores.

Because no learned judge is ground truth, evaluation validity is also central. RewardBench and JudgeBench provide preference and objective-correctness tests for reward models and language-model judges \citep{lambert2025rewardbench,tan2024judgebench}; self-preference and model-family effects further show why a held-out judge need not be an independent quality anchor \citep{panickssery2024llm}. Appendix~\ref{app:additional-related-work} develops these connections and states the novelty boundary in greater detail.

\section{Setup}
\label{sec:setup}

Let $x\in\calX$ be a prompt and $a\in\calA_x$ a candidate response. Judge $j\in\{1,\ldots,J\}$ assigns a scalar score $s_j(x,a)$. Since independently trained judges can differ in scale and offset, we map their outputs to a common audited scale,
\begin{equation}
r_j(x,a)=\frac{s_j(x,a)-\mu_j}{\tau_j},
\qquad \tau_j>0,
\label{eq:calibration}
\end{equation}
where the calibration split is disjoint from search, training, and final evaluation. A rank transform alone does not justify additive comparisons across judges; when raw outputs are not interval-scaled, the mapping must be learned against a common calibration target, for example through held-out isotonic calibration.

For a fixed prompt, write the vector of calibrated scores as
\begin{equation}
\mathbf r(x,a)=q(x,a)\one+\mathbf b(x)+\mathbf e(x,a).
\label{eq:score-decomp}
\end{equation}
Here $q(x,a)$ is the target quality, $\mathbf b(x)$ contains judge offsets for the prompt, and $\mathbf e(x,a)$ is the response-dependent residual error. Expectations below are over a declared candidate distribution $a\sim\nu_x$, conditional on $x$. The offsets absorb the conditional error means, so $\E_{\nu_x}[\mathbf e]=\mathbf0$. A shared response-dependent failure is part of $\mathbf e$, not $\mathbf b$: unlike a prompt-constant offset, it can change which candidate wins the search.

For weights $\mathbf w$ satisfying $\one^\top\mathbf w=1$, the aggregate score and error are
\begin{equation}
R_{\mathbf w}(x,a)=\mathbf w^\top\mathbf r(x,a),
\qquad
z_{\mathbf w}(x,a)=\mathbf w^\top\mathbf e(x,a).
\label{eq:ensemble}
\end{equation}
The main text uses the uniform mean $\mathbf w=\one/J$, denoted by $R$, $\bar b$, and $\bar e$. Let $\Sigma_x=\Cov_{\nu_x}(\mathbf e)$ and let
\begin{equation}
P=I-\frac{1}{J}\one\one^\top
\label{eq:projector}
\end{equation}
be the orthogonal projector away from the aggregation direction. These two projections, $\one\one^\top/J$ and $P$, organize the remainder of the analysis. We construct preference data using the calibrated disagreement-selection
DPO arm described in Algorithm~\ref{alg:selection}.

\section{Covariance Geometry Under Finite Search}
\label{sec:theory}

\subsection{The aggregation direction}

We first retain the equal-variance formulation because it gives a transparent scalar summary.

\begin{assumption}[Calibrated equal-variance judge errors]
\label{ass:equal}
For a fixed prompt $x$, $\E[e_j]=0$, $0<\Var(e_j)=\sigma^2<\infty$, and $\rho_{jk}=\Cov(e_j,e_k)/\sigma^2$. The average pairwise correlation is
\[
\rho=\frac{1}{J(J-1)}\sum_{j\ne k}\rho_{jk},
\qquad -\frac{1}{J-1}\leq\rho\leq1.
\]
\end{assumption}

\begin{lemma}[Variance of the ensemble-mean error]
\label{lem:variance}
Under Assumption~\ref{ass:equal},
\begin{equation}
\Var(\bar e)=\sigma^2\left(\frac{1}{J}+\frac{J-1}{J}\rho\right).
\label{eq:variance}
\end{equation}
\end{lemma}

Equation~\eqref{eq:variance} is the continuous-score analogue of the classical design effect \citep{kish1965survey}. Defining
\begin{equation}
J_{\mathrm{eff}}=\frac{J}{1+(J-1)\rho}
\label{eq:effective-j}
\end{equation}
gives $\Var(\bar e)=\sigma^2/J_{\mathrm{eff}}$, directly connecting the analysis to recent effective-vote measurements for discrete judge panels \citep{kohli2026nine}. The identity is exact but is not itself our novelty claim. In the heteroskedastic case, the corresponding quantity is simply
\begin{equation}
\Var(\bar e)=\frac{1}{J^2}\one^\top\Sigma_x\one.
\label{eq:matrix-var}
\end{equation}
For a heterogeneous panel, we use the variance-equivalent effective size
\begin{equation}
\bar\sigma_x^2=\frac{1}{J}\operatorname{tr}(\Sigma_x),
\qquad
J_{\mathrm{eff}}^{\mathrm{het}}
=\frac{\bar\sigma_x^2}{\Var(\bar e)}
=\frac{J\operatorname{tr}(\Sigma_x)}{\one^\top\Sigma_x\one}.
\label{eq:heterogeneous-effective-j}
\end{equation}
This quantity equals Equation~\eqref{eq:effective-j} under equal marginal
variance.  In the experiments,
$\widehat J_{\mathrm{eff}}^{\mathrm{het}}$ is computed from the
calibration-split residual covariance exactly via
Equation~\eqref{eq:heterogeneous-effective-j}---the implementation's
$J/(1+(J-1)\bar c_{\rm off}/\bar c_{\rm diag})$, with mean off-diagonal and
diagonal entries of $\widehat\Sigma_{\rm cal}$, is algebraically
identical---as a single point estimate; no bootstrap interval is reported
for it.

\subsection{Disagreement is the orthogonal component}

Define calibrated cross-judge disagreement by
\begin{equation}
\begin{split}
D(x,a)
&=\frac{1}{J}\sum_{j=1}^J\left(r_j(x,a)-R(x,a)\right)^2 \\
&=\frac{1}{J}\norm{P\mathbf r(x,a)}^2.
\end{split}
\label{eq:disagreement}
\end{equation}

\begin{proposition}[Exact projector decomposition]
\label{prop:disagreement}
Under the setup in Section~\ref{sec:setup},
\begin{equation}
\E[D(x,a)]
=\frac{1}{J}\norm{P\mathbf b(x)}^2
+\frac{1}{J}\operatorname{tr}(P\Sigma_x).
\label{eq:projector-disagreement}
\end{equation}
If the calibrated prompt offsets are common across judges and Assumption~\ref{ass:equal} holds, this reduces to
\begin{equation}
\E[D(x,a)]
=\sigma^2\frac{J-1}{J}(1-\rho).
\label{eq:disagreement-expectation}
\end{equation}
\end{proposition}

The first term in Equation~\eqref{eq:projector-disagreement} is necessary: global score calibration does not automatically remove judge-specific prompt offsets. Once that term is controlled, disagreement measures energy orthogonal to the ensemble mean. By contrast, Equation~\eqref{eq:matrix-var} measures energy parallel to the mean. In the exchangeable positive-correlation model $e_j=c+u_j$, suppose the centered shared component $c$ and centered judge-specific components $u_j$ are mutually independent. The two quantities become
\begin{equation}
\begin{aligned}
\Var(\bar e)&=\Var(c)+\frac{\Var(u_j)}{J},\\
\E[D]&=\frac{J-1}{J}\Var(u_j).
\end{aligned}
\label{eq:common-idio}
\end{equation}
Thus a shared error can produce low disagreement and high ensemble risk, while decorrelated judge-specific errors can produce high disagreement and low ensemble risk.

The separation creates an identification boundary before any tail assumption or search analysis enters.

\begin{proposition}[Identification limit]
\label{prop:nonidentifiability}
For any scalar function $h(x,a)$ with $\E_{\nu_x}[h]=0$, define
\begin{equation}
q^{(h)}=q+h,
\qquad
\mathbf e^{(h)}=\mathbf e-h\one.
\label{eq:observational-equivalence}
\end{equation}
The scores and the centering convention are unchanged, while
\begin{equation}
P\mathbf e^{(h)}=P\mathbf e,
\qquad
\bar e^{(h)}=\bar e-h.
\label{eq:nonidentifiability}
\end{equation}
Consequently, internal judge scores and their disagreement cannot identify response-dependent common-mode error without an external anchor or additional structural assumptions.
\end{proposition}

\begin{figure*}[t]
    \centering
    \includegraphics[width=0.9\textwidth]{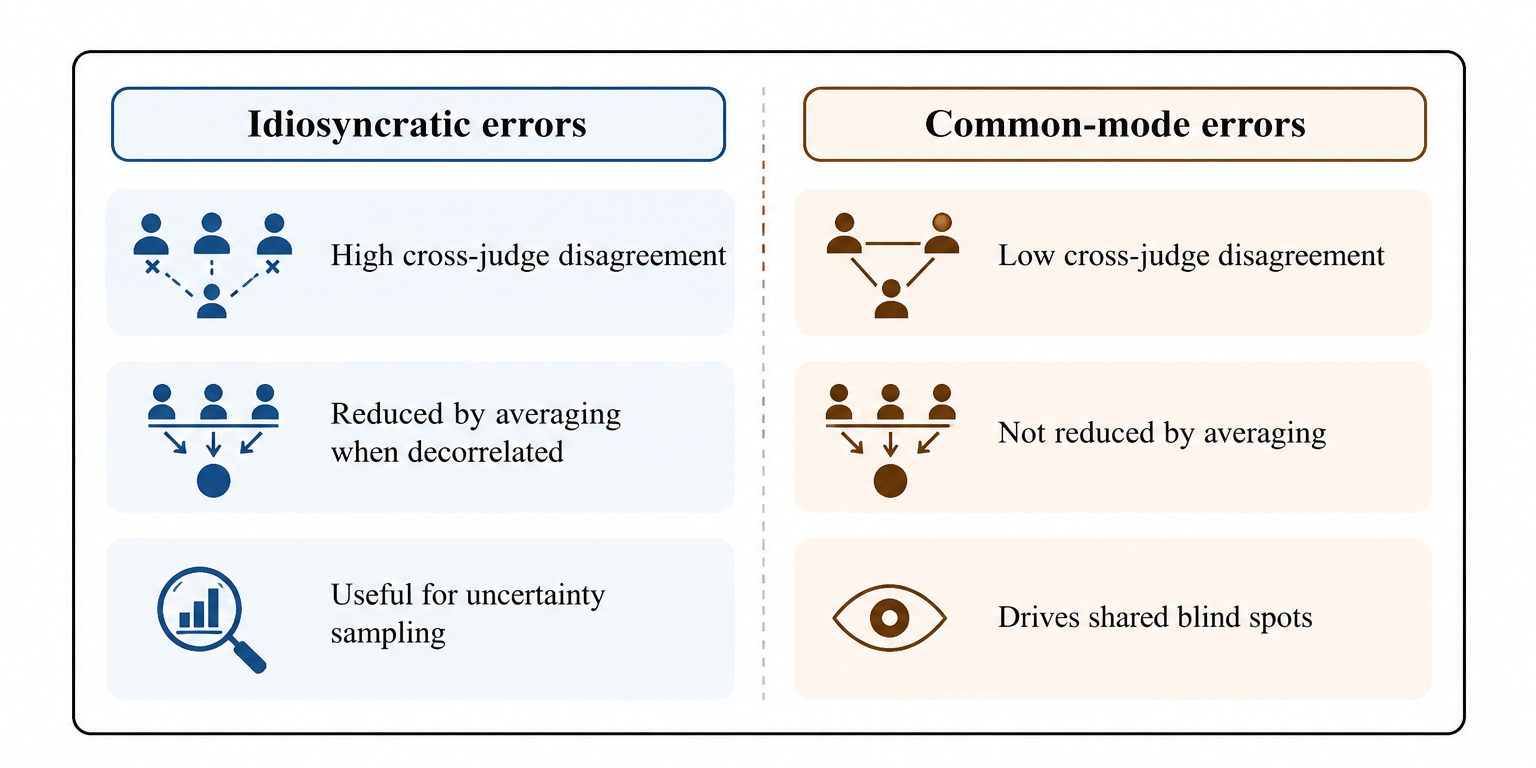}
    \caption{Conceptual distinction between idiosyncratic and common-mode judge errors. Idiosyncratic errors generate disagreement and can be reduced through averaging when sufficiently decorrelated. Common-mode errors are shared across judges, remain after averaging, and can induce reward hacking despite apparent consensus. The projector $P$ measures the former, while the aggregation direction $\one$ measures the latter.}
    \label{fig:error_modes}
    \label{fig:error-modes}
\end{figure*}
Figure~\ref{fig:error_modes} illustrates the distinction between idiosyncratic disagreement and common-mode judge error. The former is captured by the projector $P$, whereas the latter lies along the aggregation direction $\one$ and therefore survives averaging.

\subsection{What finite search can exploit}

The covariance identity controls a second moment, but reward hacking is a selection problem. For candidates $a_1,\ldots,a_K$, define
\begin{equation}
\begin{aligned}
k_R &\in \arg\max_{k\leq K} R(x,a_k),\\
k_q &\in \arg\max_{k\leq K} q(x,a_k).
\end{aligned}
\label{eq:selected-indices}
\end{equation}

\begin{assumption}[Joint sub-Gaussian search errors]
\label{ass:search}
For a fixed prompt $x$, candidates are sampled from the declared search distribution. Their centered error vectors satisfy, for every $\mathbf t\in\R^J$,
\begin{equation}
\E\exp(\mathbf t^\top\mathbf e)
\leq\exp\!\left(\frac{1}{2}\mathbf t^\top\Gamma_x\mathbf t\right)
\label{eq:joint-subg}
\end{equation}
for a positive semidefinite proxy matrix $\Gamma_x$. For the uniform ensemble, let $v_x=J^{-2}\one^\top\Gamma_x\one$. Candidate errors share this conditional marginal bound; independence across candidates is sufficient but not required by the union-bound argument.
\end{assumption}

\begin{theorem}[Selected overstatement and target-quality regret]
\label{thm:search-bound}
Under Assumption~\ref{ass:search}, the response selected by the ensemble satisfies
\begin{equation}
\E\left[R(x,a_{k_R})-q(x,a_{k_R})\right]
\leq \bar b(x)+\sqrt{2v_x\log K}.
\label{eq:expected-bound}
\end{equation}
With probability at least $1-\delta$,
\begin{equation}
R(x,a_{k_R})-q(x,a_{k_R})
\leq \bar b(x)+\sqrt{2v_x\log(K/\delta)}.
\label{eq:high-prob-bound}
\end{equation}
The loss in target quality relative to the best searched candidate obeys
\begin{align}
\E\left[q(x,a_{k_q})-q(x,a_{k_R})\right]
&\leq2\sqrt{2v_x\log K},
\label{eq:expected-regret}\\
q(x,a_{k_q})-q(x,a_{k_R})
&\leq2\sqrt{2v_x\log(2K/\delta)}
\label{eq:hp-regret}
\end{align}
with probability at least $1-\delta$.
\end{theorem}

The constant prompt offset $\bar b(x)$ affects absolute overstatement but cancels from the within-prompt regret. The exploitable term is the response-dependent error projected onto the ensemble direction. For general sub-Gaussian errors, Theorem~\ref{thm:search-bound} is an upper bound rather than an equality. If the projected candidate errors are independent Gaussian variables with variance $v_x$, classical extreme-value theory gives $\E\max_k\bar e_k=\sqrt{2v_x\log K}(1+o(1))$, so the order is asymptotically tight \citep{leadbetter1983extremes}.

\begin{corollary}[Error diversity]
\label{cor:diversity}
Under Assumption~\ref{ass:equal} and a Gaussian proxy, or a sub-Gaussian proxy with matching covariance, the variance-driven terms in Theorem~\ref{thm:search-bound} decrease as $d=1-\rho$ increases, holding $J$, $K$, and $\sigma$ fixed. Diversity does not remove a shared response-dependent component or the prompt offset.
\end{corollary}

The same theorem applies to any fixed weights $\mathbf w$ after replacing $v_x$ by $\mathbf w^\top\Gamma_x\mathbf w$. When $\Gamma_x$ is positive definite and unconstrained signed weights are allowed, the minimum-variance weights are proportional to $\Gamma_x^{-1}\one$. Nonnegative weights require the corresponding simplex-constrained quadratic program. This observation motivates covariance-aware ensemble construction but does not address unobserved bias by itself.

The fixed-distribution statement also has a finite adaptive counterpart. It is deliberately conditional: adaptive search is safe only if calibration continues to hold after conditioning on the search history.

\begin{corollary}[Predictable finite adaptation]
\label{cor:adaptive-search}
Let $\mathcal{F}_{k-1}$ contain the prompts, candidates, scores, and random choices observed before proposal $k$. The proposal distribution for $a_k$ may be $\mathcal{F}_{k-1}$-measurable. If, for every $k\leq K$ and $\lambda\in\R$,
\begin{equation}
\begin{aligned}
\E\!\left[\exp(\lambda z_k)\mid\mathcal{F}_{k-1}\right]
&\leq \exp(\lambda^2v_x/2),\\
z_k&=\bar e(x,a_k),
\end{aligned}
\label{eq:adaptive-subg}
\end{equation}
then all four bounds in Theorem~\ref{thm:search-bound} hold for selection from the $K$ adaptively proposed candidates.
\end{corollary}

\subsection{Noise in the quality anchor}

Proposition~\ref{prop:nonidentifiability} makes an external quality signal necessary. Yet the covariance in Equation~\eqref{eq:matrix-var} is defined relative to the target quality, which is rarely observed. Let a held-out quality proxy be $\widetilde q=q+\eta$, and define proxy-relative errors $\widetilde{\mathbf e}=\mathbf r-\widetilde q\one-\mathbf b$.

\begin{proposition}[Quality-anchor contamination]
\label{prop:proxy-contamination}
If $\eta$ is centered, independent of $\mathbf e$, and has variance $\tau^2$, then
\begin{equation}
\Cov(\widetilde{\mathbf e})
=\Sigma_x+\tau^2\one\one^\top.
\label{eq:proxy-contamination}
\end{equation}
Consequently, the uniform-ensemble variance is inflated by $\tau^2$, whereas $P\widetilde{\mathbf e}=P\mathbf e$ and disagreement is unchanged.
\end{proposition}

The contamination therefore looks exactly like an error shared by all judges. A single stronger judge can still define a useful proxy-relative audit, but it cannot identify true common-mode risk without assumptions about its own error.

\begin{corollary}[Two-anchor decontamination]
\label{cor:two-anchor}
Suppose $\widetilde q^{(1)}=q+\eta_1$ and $\widetilde q^{(2)}=q+\eta_2$, where the anchor errors are centered and conditionally independent of each other and of $\mathbf e$. If $\widetilde{\mathbf e}^{(m)}=\mathbf r-\widetilde q^{(m)}\one-\mathbf b$, then
\begin{equation}
\Cov\!\left(\widetilde{\mathbf e}^{(1)},
\widetilde{\mathbf e}^{(2)}\right)=\Sigma_x.
\label{eq:two-anchor}
\end{equation}
\end{corollary}

This cross-covariance construction is most credible when the anchors arise from different mechanisms, such as an exact verifier and a held-out judge, rather than two closely related language models.

\subsection{From estimated variance to a valid bounded tail}
\label{sec:bounded-tail-certificate}

The covariance analysis identifies the direction that search can exploit, but
covariance alone does not certify a sub-Gaussian proxy. We close this gap on a
declared bounded score scale: one split fits the aggregator, a disjoint
certification split estimates its residual variance; pairing two candidates
from the same prompt removes prompt-level offsets, two quality anchors remove
the rank-one contamination above under an explicit cross-orthogonality
condition, and Bernstein's inequality converts the variance upper bound and
the declared range into a valid search tail.

Fix a predeclared task--generator stratum $s$ and an aggregation rule $A$ that
is fixed independently of the certification split. For a random prompt $X$ in
this stratum and candidate $a\sim\nu_X$, write
\begin{equation}
\begin{aligned}
Y(X,a)&=A(\mathbf r(X,a))-q(X,a),\\
Z(X,a)&=Y(X,a)-\E[Y(X,a)\mid X].
\end{aligned}
\label{eq:bounded-centered-error}
\end{equation}
Thus $\E[Z\mid X]=0$, and the prompt-constant offset cancels from
within-prompt selection.

\begin{theorem}[Bounded two-anchor Bernstein certificate]
\label{thm:bounded-anchor-search}
Let $(X_i,a_i,a_i')_{i=1}^m$ be independent certification triples ($X_i$
from stratum $s$; $a_i,a_i'$ conditionally independent draws from
$\nu_{X_i}$), independent of the subsequent search sample. For anchors
$\widetilde q^{(\ell)}=q+\eta^{(\ell)}$, $\ell\in\{1,2\}$, define
$U^{(\ell)}(X,a)=A(\mathbf r(X,a))-\widetilde q^{(\ell)}(X,a)$
and
$\Delta_i^{(\ell)}
=U^{(\ell)}(X_i,a_i)-U^{(\ell)}(X_i,a_i')$.
Assume, conditional on $X$, that each anchor error is centered, each is
uncorrelated with $Z$, and the two anchor errors are mutually uncorrelated:
\begin{equation}
\begin{aligned}
\E[\eta^{(\ell)}\mid X]&=0,
&\E[Z\eta^{(\ell)}\mid X]&=0,\\
\E[\eta^{(1)}\eta^{(2)}\mid X]&=0.&&
\end{aligned}
\label{eq:anchor-cross-orthogonality}
\end{equation}
Suppose the declared score and anchor ranges imply
$|\Delta_i^{(\ell)}|\leq C_{\ell,s}$ and $|Z|\leq c_s$ almost surely. Set
\begin{equation}
\begin{aligned}
\widehat v_s
&=\frac{1}{2m}\sum_{i=1}^m
\Delta_i^{(1)}\Delta_i^{(2)},\\
v_s^{\mathrm U}
&=\max\!\left\{0,\widehat v_s
+C_{1,s}C_{2,s}
\sqrt{\frac{\log(1/\delta_{\rm est})}{2m}}\right\}.
\end{aligned}
\label{eq:bounded-anchor-variance-ucb}
\end{equation}
Then, with probability at least $1-\delta_{\rm est}$ over the certification
sample,
\begin{equation}
\Var(Z)\leq v_s^{\mathrm U}.
\label{eq:variance-ucb-event}
\end{equation}
For a new random prompt from $s$ and any $K$ searched candidates having the
same marginal candidate law, define
\begin{equation}
B_s(K,\delta;v)
=\sqrt{2v\log(K/\delta)}
+\frac{c_s}{3}\log(K/\delta)
\label{eq:bernstein-search-radii}
\end{equation}
and $B_s^{\pm}(K,\delta;v)=B_s(2K,\delta;v)$.
If $k_A$ maximizes the aggregate score and $k_q$ maximizes target quality,
then, jointly over certification and search, with probability at least
$1-\delta_{\rm est}-\delta_{\rm search}$,
\begin{align}
Z(X,a_{k_A})
&\leq B_s(K,\delta_{\rm search};v_s^{\mathrm U}),
\label{eq:bounded-selected-error}\\
q(X,a_{k_q})-q(X,a_{k_A})
&\leq 2B_s^{\pm}(K,\delta_{\rm search};v_s^{\mathrm U}).
\label{eq:bounded-search-regret}
\end{align}
The probability statement is marginal over prompts within the predeclared
stratum. It becomes prompt-conditional only when the assumptions and the
variance upper bound are established separately for that prompt.
\end{theorem}

The theorem does not require Gaussian errors and does not equate covariance
with a sub-Gaussian proxy; its linear term records the price of a bounded but
otherwise unknown tail. A tighter Bennett inversion, the
declared-constant conventions, and a Bonferroni-simultaneous version
appear with the proof in Appendix~\ref{app:proofs}; other
proofs are in Appendix~\ref{app:proofs}. The assumptions, failure modes, and claim boundaries underlying these conclusions are detailed in Appendix~\ref{app:claim-boundaries}.

\section{Empirical Audits}
\label{sec:experiments}

\subsection{Controlled Gaussian stress test}

Before using learned evaluators, we test the implementation in the model where the order of the maximum-error envelope is known to be tight. The parameter grid is

\begin{equation}
\begin{aligned}
J &\in \{1,2,4,8\},\\
K &\in \{2,4,8,16,32\},\\
\rho &\in \{0,.25,.50,.65,.75,.90\}.
\end{aligned}
\label{eq:synthetic-grid}
\end{equation}
For each setting, we generate $50{,}000$ prompts with $q_k\sim\mathcal N(0,1)$ and
\begin{equation}
e_{jk}=\sqrt{\rho}\,c_k+\sqrt{1-\rho}\,u_{jk},
\qquad c_k,u_{jk}\stackrel{\mathrm{iid}}{\sim}\mathcal N(0,1).
\label{eq:synthetic-model}
\end{equation}
The fixed seed is 20260802, and nested candidate prefixes are reused across $K$. Across the 120 cells no Monte Carlo mean exceeds its population upper bound (largest observed-to-bound ratio $0.787$ for the envelope, $0.557$ for overstatement, $0.116$ for regret; every MC standard error below $0.0042$). Figure~\ref{fig:synthetic-stress} also checks the quality-anchor algebra: one noisy anchor inflates the estimated projected variance by $\approx\tau^2$ while the two-anchor cross-covariance stays within $0.002$ of the true $0.625$ at all noise levels. These verify the code path and the rank-one contamination identity under their stated model---not that real judge errors are Gaussian or real anchors independent (full outputs: Appendix~\ref{app:synthetic-stress}).

\subsection{Multi-family finite-search audit}
\label{sec:multifamily}
\label{sec:bound-validation}

The archived multi-family audit (five models, three families, two
generators, one panel per row; Appendix~\ref{app:multifamily-full},
Table~\ref{tab:multifamily}) retains three observations: the projected
scale falls with panel size but
$\widehat J_{\rm eff}^{\rm het}\approx1.5$ at $J{=}4$; overstatement falls
with $J$ faster than regret improves (descriptive---the records audit
below supplies the paired version); the $0.81$--$0.99$ plug-in pass
fractions are not coverage. Both $J{=}2$ panels are within-DeBERTa, so no
same- versus cross-family conclusion is supported. The complementary two-judge error-envelope audit is reported in Appendix~\ref{app:bound-table}.

\subsection{Anchor sensitivity and verifier-backed evaluation (summary)}
\label{sec:targetquality}

Because the audits above measure error relative to a held-out learned judge, we additionally examine the sensitivity of the results to the choice of quality anchor by comparing two held-out anchors from different model families (Appendix~\ref{app:anchor-sensitivity}). The anchors exhibit Spearman correlation $0.42$ on GSM8K but $-0.21$ on open-ended prompts, with corresponding anchor-dependent regret, whereas HumanEval provides a degenerate verifiable anchor. The empirical validation of the two-anchor identity and its associated UCB is therefore confined to the controlled synthetic experiment.
\begin{table*}[t]
\centering
\small
\begin{tabular}{lcc}
\toprule
Aggregator & selected quality & target regret \\
\midrule
best\_single & 0.075 [0.033, 0.125] & 0.292 [0.217, 0.375] \\
care & 0.092 [0.042, 0.142] & 0.275 [0.200, 0.358] \\
covariance\_weighted & 0.108 [0.058, 0.167] & 0.258 [0.183, 0.342] \\
minimum & 0.108 [0.058, 0.167] & 0.258 [0.183, 0.342] \\
uncertainty\_penalized & 0.108 [0.058, 0.167] & 0.258 [0.183, 0.333] \\
uniform\_mean & 0.100 [0.050, 0.158] & 0.267 [0.192, 0.350] \\
\bottomrule
\end{tabular}

\caption{Aggregation comparison on the largest eligible panel ($J{=}5$) and
search budget ($K{=}32$): mean selected target quality and target regret
with prompt-paired bootstrap $95\%$ intervals over the $120$ locked test
prompts. Best-single, penalty, covariance weights, and CARE
hyperparameters are chosen without test access. }
\label{tab:aggregation-baselines}
\end{table*}

\subsection{All-subset and aggregation audit}
\label{sec:all-subset-audit}

We freeze five eligible judges and enumerate every nonempty subset
($31$ panels; held-out anchors excluded). Every panel receives the same
tensors ($320$ GSM8K prompts, $K_{\max}{=}32$ nested candidates from a
frozen SmolLM2-360M sampler; input SHA-256 digests released);
$K\in\{2,4,8,16,32\}$ uses nested prefixes, so all contrasts are
prompt-paired. Disjoint fit/certification/test splits ($120/80/120$)
separate learned components, the variance UCB, and final metrics. Six
predeclared rules are compared (best-single, uniform mean, minimum,
uncertainty-penalized, Ledoit--Wolf simplex covariance weighting, CARE-SVD
at the pinned official revision; CARE marked not applicable on $J{<}3$
panels rather than silently replaced). Intervals resample prompts; method
contrasts subtract the within-prompt uniform mean; panel-size contrasts
average \emph{all} $\binom{5}{j}$ panels of a size within prompt before
subtracting the singleton average, so ensemble size is no longer
confounded with judge identity---the confound that made the archived
$J{=}1$ vs.\ $J{=}4$ comparison descriptive. Additional experimental-design and reporting requirements, including cross-family evaluation and acquisition diagnostics, are provided in Appendix~\ref{app:experimental-design}. Scope: the generator solves only $3\%$ of items ($41\%$ of
prompts have a correct candidate)---low-prevalence contrasts.

The paired results are informative in both directions
(Table~\ref{tab:aggregation-baselines}). \emph{No aggregation rule
separates from the uniform mean on selected target quality} at $J{=}5$,
$K{=}32$ (best-single $-0.025$ $[-0.067,0.008]$; CARE $-0.008$
$[-0.050,0.033]$; the weighted rules $+0.008$ $[-0.017,0.042]$). The rules
separate on \emph{proxy inflation}: best-single overstates the selected
centered score by $+0.104$ $[0.067,0.144]$ versus the mean; covariance
weighting is the only rule not worse ($-0.018$ $[-0.049,0.009]$).
Ensembling itself helps: $J{=}5$ beats singletons on selected quality
($+0.028$ $[0.003,0.058]$) and cuts centered overstatement ($-0.107$
$[-0.139,-0.080]$; covariance $-0.126$ $[-0.162,-0.094]$)---the
common-mode account on real records with prompt-paired uncertainty.

\subsection{Real-task bounded certificate (summary)}
\label{sec:real-anchor-certificate}

On the one predeclared non-degenerate task (GSM8K: exact match is the target and anchor~1, so $\eta_1\equiv0$; GRM-Llama3 is anchor~2), Theorem~\ref{thm:bounded-anchor-search} is \emph{valid but uninformative across all $855$ cells}: at $m{=}80$, the estimation correction ($\approx0.61$) dominates $\widehat v_s$ (median $0.017$), every radius reaches its range cap, and the $1.0$ pass fractions are trivial and therefore do not constitute coverage. Achieving $\widehat v_s$-scale precision would require approximately $10^{5}$ pairs (Appendix~\ref{app:anchor-ucb-cells}). A three-seed DPO pilot (Appendix~\ref{sec:selection-result}), while underpowered, provides no statistically reliable evidence for either superiority or equivalence. The full calibrated disagreement-selection protocol, including the matching controls and acquisition procedure, is provided in Appendix~\ref{app:dpo-protocol}. A detailed analysis of the three-seed DPO pilot (Appendix~\ref{app:selection-analysis}) further delineates the interpretation and limitations of this result; in particular, the pilot neither establishes a benefit nor rules out a practically meaningful effect.
The corresponding seed-level confidence interval and exact randomization-test calculations are reported in Appendix~\ref{app:statistics}. For visualization, see Appendix~\ref{app:supp}.

\section{Conclusion and Implications}

Evaluator ensembles mitigate reward hacking only along directions in
which their errors cancel: covariance projected onto the aggregation
direction controls the error exposed to finite search; disagreement
measures an orthogonal component. This yields overstatement and regret
bounds (incl.\ conditionally calibrated adaptive proposals); only the
maximum-error envelope has a general Gaussian tightness result; a noisy
proxy adds a rank-one common mode, removable by two anchors only under
explicit conditional independence. Empirically, ensembling suppresses
proxy inflation and modestly improves selected quality, no weighting rule
beats the plain mean, and the real-task certificate is valid but
uninformative at audit scale. Panels, budgets, calibration, and external
quality evidence must be audited jointly.

\section*{Limitations}
\label{sec:limitations}

The Gaussian-model search guarantees are conditional on a joint sub-Gaussian proxy, or on history-conditional control for predictable adaptation, and covariance alone does not imply either tail condition. The bounded Bernstein certificate (Theorem~\ref{thm:bounded-anchor-search}) removes the earlier Gaussian bridge, but it does not make the anchor assumptions automatic: its validity requires a predeclared bounded scale, a fit split independent of certification, and centered cross-orthogonal anchor errors; the main statement is marginal over prompts in a declared stratum, not prompt-conditional; the linear Bernstein term can make the certificate conservative, especially after simultaneous correction; and learned anchors from different families remain a sensitivity analysis unless their error relationship has independent design-based support. The archived real-model experiments use covariance plug-ins rather than certified proxy matrices. The multi-family audit spans five eligible ensemble judges from three families plus a held-out Llama-family proxy, two small generators, and $200$ locked test prompts, with one executed panel per row rather than all subsets; scaling to frontier-size generators, larger panels, and non-English tasks is unestablished. The expectation-scale benchmark pass fraction is as low as $0.81$, but this descriptive quantity is not nominal coverage because the real-model audit does not certify a prompt-conditional sub-Gaussian proxy; real-data nominal coverage of Equation~\eqref{eq:high-prob-bound} is unestablished. Both executed $J{=}2$ panels are within-DeBERTa pairs, so no same- versus cross-family conclusion is available from this run, and no paired $J{=}1$ versus $J{=}4$ contrast, frozen common-mode thresholds, exact checkpoint revision hashes, or per-prompt records are contained in the archived summaries. Direct aggregation baselines (best single judge, minimum, uncertainty-penalized, simplex-constrained covariance weighting, and CARE at the pinned official revision) are now executed on the all-subset records audit (\S\ref{sec:all-subset-audit}); that audit is single-task (GSM8K), single-generator, and low-prevalence, so its contrasts do not generalize beyond that scope, and the real-task certificate it feeds is uninformative at $m{=}80$ (\S\ref{sec:real-anchor-certificate}). The archived $z$-scored runs were unbounded without a predeclared clipping rule, so no certificate value is derived from them. The measured covariances remain proxy-relative (Proposition~\ref{prop:proxy-contamination}); the two-anchor validation mitigates but does not eliminate this, and it shows the anchors themselves disagree beyond sign outside verifiable domains---GSM8K supplies the only non-degenerate verifiable anchor here, since the $1.5$B generator passes zero HumanEval suites. The DPO study contains only three paired training seeds and one reported external judge; it is underpowered for equivalence, and no human evaluation was available. A power-determined multi-seed study with a predeclared smallest effect of interest remains necessary. Corollary~\ref{cor:adaptive-search} covers finite predictable proposals only when conditional calibration is justified; it does not cover fully co-evolving policy training in which that condition can fail.

\section{Ethics Statement}

This work studies failures of automated evaluation and does not introduce human-subject data. Nevertheless, reward-hacking diagnostics can be dual use: the same analyses that identify evaluator weaknesses could help an adversary target them. 

\bibliography{references}

\appendix

\section{Proofs}
\label{app:proofs}

\paragraph {Proof of Lemma~\ref{lem:variance}}

Using $\bar e=J^{-1}\sum_j e_j$,
\begin{align*}
\Var(\bar e)
&=\frac{1}{J^2}\left(\sum_j\Var(e_j)+\sum_{j\ne k}\Cov(e_j,e_k)\right)\\
&=\frac{1}{J^2}\left(J\sigma^2+J(J-1)\rho\sigma^2\right),
\end{align*}
which is Equation~\eqref{eq:variance}. Positive semidefiniteness implies $\rho\geq-1/(J-1)$. At $\rho=1$, averaging provides no variance reduction; at $\rho=0$, the standard deviation falls as $1/\sqrt{J}$; and at the negative endpoint, the equal-variance errors cancel in the mean.

\paragraph{Proof of Proposition~\ref{prop:disagreement}}

Because $P\one=0$, Equation~\eqref{eq:score-decomp} gives $P\mathbf r=P\mathbf b+P\mathbf e$. Since $\E[\mathbf e]=0$,
\begin{align*}
\E[D]
&=\frac{1}{J}\E\norm{P\mathbf b+P\mathbf e}^2\\
&=\frac{1}{J}\norm{P\mathbf b}^2
+\frac{1}{J}\E[\mathbf e^\top P\mathbf e]\\
&=\frac{1}{J}\norm{P\mathbf b}^2
+\frac{1}{J}\operatorname{tr}(P\Sigma_x).
\end{align*}
If $P\mathbf b=0$ and all marginal variances equal $\sigma^2$, then
\begin{align*}
\frac{1}{J}\operatorname{tr}(P\Sigma_x)
&=\frac{1}{J}\operatorname{tr}(\Sigma_x)
-\frac{1}{J^2}\one^\top\Sigma_x\one\\
&=\sigma^2-\Var(\bar e)\\
&=\sigma^2\frac{J-1}{J}(1-\rho).
\end{align*}

\subsection{Proof of Proposition~\ref{prop:nonidentifiability}}

Substitution gives
\[
q^{(h)}\one+\mathbf b+\mathbf e^{(h)}
=(q+h)\one+\mathbf b+\mathbf e-h\one
=\mathbf r.
\]
Moreover, $\E[\mathbf e^{(h)}]=\mathbf0$ because both $\mathbf e$ and $h$ are centered under $\nu_x$. Finally, $P\one=0$ gives $P\mathbf e^{(h)}=P\mathbf e$, whereas uniform averaging gives $\bar e^{(h)}=\bar e-h$. The observations and every statistic computed solely from them are therefore unchanged even though the common-mode error differs.

\paragraph{Proof of Theorem~\ref{thm:search-bound}}

Let $z_k=\bar e(x,a_k)$. Assumption~\ref{ass:search} with $\mathbf t=\lambda\one/J$ gives $\E\exp(\lambda z_k)\leq\exp(\lambda^2v_x/2)$. For any $\lambda>0$,
\begin{align*}
\E\max_k z_k
&\leq\frac{1}{\lambda}\log\E\exp\!\left(\lambda\max_kz_k\right)\\
&\leq\frac{1}{\lambda}\log\sum_{k=1}^K\E\exp(\lambda z_k)\\
&\leq\frac{\log K}{\lambda}+\frac{\lambda v_x}{2}.
\end{align*}
Optimizing at $\lambda=\sqrt{2\log K/v_x}$ yields $\E\max_kz_k\leq\sqrt{2v_x\log K}$. Since
\[
R(x,a_{k_R})-q(x,a_{k_R})
=\bar b(x)+z_{k_R}
\leq\bar b(x)+\max_kz_k,
\]
Equation~\eqref{eq:expected-bound} follows. The sub-Gaussian tail and a union bound give
\[
\Prb\!\left(\max_kz_k>t\right)
\leq K\exp\!\left(-\frac{t^2}{2v_x}\right),
\]
which proves Equation~\eqref{eq:high-prob-bound}.

For regret, the definition of $k_R$ gives
\[
q_{k_R}+z_{k_R}\geq q_{k_q}+z_{k_q},
\]
because the prompt offset is common to all candidates. Therefore
\[
q_{k_q}-q_{k_R}
\leq z_{k_R}-z_{k_q}
\leq\max_kz_k-\min_kz_k.
\]
Applying the expected maximum bound to both $z_k$ and $-z_k$ proves Equation~\eqref{eq:expected-regret}. A union bound over both tails of all $K$ variables gives $\max_k|z_k|\leq\sqrt{2v_x\log(2K/\delta)}$ with probability at least $1-\delta$, and the range is at most twice this value, proving Equation~\eqref{eq:hp-regret}.

\paragraph{Proof of Corollary~\ref{cor:adaptive-search}}

Taking expectations in Equation~\eqref{eq:adaptive-subg} gives

\[
\begin{aligned}
\E\exp(\lambda z_k)
&= \E\!\left[
    \E\{\exp(\lambda z_k)\mid\mathcal F_{k-1}\}
   \right] \\
&\leq \exp(\lambda^2v_x/2).
\end{aligned}
\]
for every adaptively generated candidate. The log-sum-exp argument in the proof of Theorem~\ref{thm:search-bound} therefore applies without independence. Likewise,

\[
\begin{aligned}
\Prb(z_k>t)
&= \E\!\left[
    \Prb(z_k>t\mid\mathcal F_{k-1})
   \right] \\
&\leq \exp(-t^2/2v_x).
\end{aligned}
\]
and the same statement holds for $-z_k$. Union bounds over the predictable sequence establish the two high-probability conclusions. The regret comparison remains pathwise after the $K$ candidates have been generated, so its proof is unchanged.

\paragraph{Proof of Corollary~\ref{cor:diversity}}

Let $d=1-\rho$. The variance-driven expected overstatement term is
\[
B(d)=\sigma\sqrt{2\log K}
\left(\frac{1}{J}+\frac{J-1}{J}(1-d)\right)^{1/2}.
\]
At every interior point where the variance is positive,

\[
\begin{aligned}
\frac{\partial B}{\partial d}
&=
-\frac{\sigma\sqrt{2\log K}}{2}
\frac{J-1}{J} \\
&\qquad\times
\left(
\frac{1}{J}
+\frac{J-1}{J}\rho
\right)^{-1/2}
<0.
\end{aligned}
\]
The same monotonicity applies to the regret bound. Neither derivative contains the constant prompt offset, and the equal-variance parameterization holds $\sigma$ fixed.

\paragraph{Proofs of Proposition~\ref{prop:proxy-contamination} and Corollary~\ref{cor:two-anchor}}

The proxy-relative error is $\widetilde{\mathbf e}=\mathbf e-\eta\one$. Independence and centering give
\begin{align*}
\Cov(\widetilde{\mathbf e})
&=\Cov(\mathbf e)+\Var(\eta)\one\one^\top\\
&=\Sigma_x+\tau^2\one\one^\top.
\end{align*}
For uniform weights, $J^{-2}\one^\top(\tau^2\one\one^\top)\one=\tau^2$. In contrast, $P\widetilde{\mathbf e}=P\mathbf e-\eta P\one=P\mathbf e$ pointwise.

With two anchors,
\[
\widetilde{\mathbf e}^{(1)}=\mathbf e-\eta_1\one,
\qquad
\widetilde{\mathbf e}^{(2)}=\mathbf e-\eta_2\one.
\]
All mixed covariance terms vanish under the conditional independence assumptions, leaving
\[
\Cov\!\left(\widetilde{\mathbf e}^{(1)},
\widetilde{\mathbf e}^{(2)}\right)=\Cov(\mathbf e)=\Sigma_x.
\]

\paragraph{Bennett refinement, declared constants, and simultaneity}
\label{app:bounded-tail-details}

The closed-form Bernstein radius is convenient but need not be the tightest
valid use of the same variance estimate. Let
$h(u)=(1+u)\log(1+u)-u$ and
$\bar v=\min\{v_s^{\mathrm U},c_s^2\}$. The deterministic cap follows
directly from the theorem's assumption $|Z|\leq c_s$, which gives
$\operatorname{Var}(Z\mid X)\leq\E[Z^2\mid X]\leq c_s^2$; the sharper
Popoviciu cap $c_s^2/4$ would additionally require the conditional support
\emph{width} to be at most $c_s$, which the theorem does not assume, so we
do not use it. Define
\begin{equation}
T_s(K,\delta;v_s^{\mathrm U})
=\min\!\left\{
c_s,
\frac{\bar v}{c_s}
h^{-1}\!\left(
\frac{c_s^2}{\bar v}\log\frac{K}{\delta}
\right)
\right\},
\label{eq:bennett-search-radius}
\end{equation}
with $T_s=0$ when $\bar v=0$. Bennett's bounded-variance moment-generating
function bound implies that Equation~\eqref{eq:bounded-selected-error} remains
valid after replacing $B_s$ by $T_s$. If target quality has declared range
width $W_{q,s}$, Equation~\eqref{eq:bounded-search-regret} remains valid with
the tighter radius
\begin{equation}
\min\!\left\{
W_{q,s},
2T_s(2K,\delta_{\rm search};v_s^{\mathrm U})
\right\}.
\label{eq:bennett-regret-radius}
\end{equation}
We report the Bennett radius together with the closed-form Bernstein radius
and the deterministic cap. The standard inequality
$h(u)\geq u^2/[2(1+u/3)]$ shows pointwise that the Bennett inversion is no
larger than the Bernstein relaxation, while the deterministic cap holds
surely. Thus reporting the tightest of these nested bounds introduces no
post-selection or additional failure probability.

If the aggregate and each anchor are clipped to
$[0,1]$ before either tuning or certification, then
$C_{1,s}=C_{2,s}=2$. If both aggregate and target quality lie in $[0,1]$,
then $c_s=2$ is valid because centering a variable with range two produces an
absolute deviation of at most two. These constants must be declared rather
than estimated from the observed extrema.

For $M$ predeclared cells, choosing
$\delta_{\rm est}=\delta_{\rm search}=\alpha/(2M)$ and applying a union bound
makes Equations~\eqref{eq:bounded-selected-error}--\eqref{eq:bounded-search-regret}
simultaneous across all $M$ cells with family-wise probability at least
$1-\alpha$. Pointwise and simultaneous certificates should be reported in
separate columns.

\paragraph{Proof of Theorem~\ref{thm:bounded-anchor-search}}

For one certification pair, write
$U^{(\ell)}=\mu_X+Z-\eta^{(\ell)}$, where
$\mu_X=\E[Y\mid X]$. Candidate differencing removes $\mu_X$. Conditional on
$X$, independence of $a$ and $a'$ together with
Equation~\eqref{eq:anchor-cross-orthogonality} gives
\begin{equation}
\E\!\left[
\frac{1}{2}\Delta^{(1)}\Delta^{(2)}\mid X
\right]
=\Var(Z\mid X).
\end{equation}
Since $\E[Z\mid X]=0$, averaging over $X$ yields
\begin{equation}
\E\!\left[\frac{1}{2}\Delta^{(1)}\Delta^{(2)}\right]
=\E[\Var(Z\mid X)]=\Var(Z).
\end{equation}
Moreover,
$\frac{1}{2}\Delta^{(1)}\Delta^{(2)}$ lies in an interval of width
$C_{1,s}C_{2,s}$. The one-sided Hoeffding inequality
\citep{hoeffding1963probability} therefore gives
\begin{equation}
\Prb\!\left\{
\Var(Z)>\widehat v_s+C_{1,s}C_{2,s}
\sqrt{\frac{\log(1/\delta_{\rm est})}{2m}}
\right\}
\leq\delta_{\rm est},
\end{equation}
which proves Equation~\eqref{eq:variance-ucb-event}.

On this event, the classical bounded-variable moment-generating-function
bound implies, for $0\leq\lambda<3/c_s$,
\begin{equation}
\log\E[e^{\lambda Z}]
\leq
\frac{\lambda^2\Var(Z)}{2(1-\lambda c_s/3)}
\leq
\frac{\lambda^2v_s^{\mathrm U}}
{2(1-\lambda c_s/3)}.
\end{equation}
The Chernoff argument consequently yields
\begin{equation}
\Prb\!\left\{
Z>\sqrt{2v_s^{\mathrm U}t}+c_st/3
\right\}\leq e^{-t}.
\label{eq:single-bernstein-tail}
\end{equation}
A direct Bennett argument \citep{bennett1962probability} instead gives
\begin{equation}
\Prb\{Z>t\}
\leq
\exp\!\left[-\frac{\bar v}{c_s^2}
h\!\left(\frac{c_st}{\bar v}\right)\right],
\qquad 0\leq t\leq c_s,
\end{equation}
where replacing the unknown variance by its upper bound weakens the tail
bound. Inverting this expression after a union bound gives the radius in
Equation~\eqref{eq:bennett-search-radius}. Applying it to both tails gives
Equation~\eqref{eq:bennett-regret-radius}.

For the closed-form relaxation, a union bound over $K$ candidate marginals
with $t=\log(K/\delta_{\rm search})$ proves
Equation~\eqref{eq:bounded-selected-error}; independence among the searched
candidates is unnecessary for this step. Applying
Equation~\eqref{eq:single-bernstein-tail} to both $Z$ and $-Z$ with
$t=\log(2K/\delta_{\rm search})$ bounds every searched error in absolute
value by $B_s^{\pm}$. Finally, aggregate-score optimality gives
\begin{equation}
q(X,a_{k_q})-q(X,a_{k_A})
\leq Z(X,a_{k_A})-Z(X,a_{k_q}),
\end{equation}
so the two-sided event proves Equation~\eqref{eq:bounded-search-regret}.
Combining the estimation and search events by a final union bound completes
the proof.

\section{Additional Related Work}
\label{app:additional-related-work}

\paragraph{Reward overoptimization and reward hacking}

RLHF fits a learned reward model to pairwise or scalar preference judgments and then
optimizes a policy against that model as a proxy for the intended objective
\citep{christiano2017deep,stiennon2020learning,ouyang2022training}. The reward model is
estimated from a finite, imperfectly labeled sample and must evaluate a policy distribution
that changes as optimization proceeds. Its reliability is therefore not a peripheral
implementation detail: a region in which the reward model is an unreliable proxy can become
the region toward which the optimized policy moves.

\citet{gao2023scaling} operationalize this failure using a large ``gold'' reward model as a
stand-in for human judgment. Optimizing a smaller proxy through best-of-$n$ sampling or PPO
initially improves gold reward, but continued optimization eventually decreases it.
\citet{rafailov2024scaling} observe an analogous pattern for direct alignment algorithms such
as DPO, showing that the phenomenon is not specific to explicit reinforcement learning
against a scalar reward. We use \emph{reward overoptimization} for this measurable degradation
under a stronger evaluation signal. We use \emph{reward hacking} more broadly for behavior
that obtains proxy reward in a way the principal would not endorse
\citep{skalse2022defining}. Overoptimization is one empirically tractable manifestation of
that broader problem.

Our contribution is not to establish that proxy optimization can fail. Instead, we ask which
component of a learned evaluator's error remains after an ensemble is aggregated and how a
finite search budget can exploit that component. This scope matters because it separates the
existence of reward hacking, which prior work already establishes, from the narrower question
of when averaging evaluators can mitigate it.

\paragraph{Reward-model ensembles and uncertainty}

The most direct ensemble mitigation replaces a single reward model with several members and
aggregates their predictions. \citet{coste2024ensembles} compare mean and worst-case
objectives with uncertainty weighting. Their experiments show that conservative aggregation
can substantially reduce overoptimization, particularly under label noise. This result also
shows why the mean should not be treated as the uniquely correct ensemble objective: different
aggregators expose different directions of the joint error and make different bias--variance
tradeoffs.

\citet{eisenstein2023helping} reach a complementary conclusion. Reward models that perform
similarly on their training distribution can be underspecified by the available preference
data and assign sharply different rewards after the policy moves off distribution. Ensembles
that vary in pretraining seed transfer better under optimization than ensembles that vary
only in fine-tuning seed on top of a shared pretrained model. Even pretraining-diverse
ensembles do not eliminate reward hacking, however, because some qualitative failures are
shared by every member. \citet{rame2024warm} study weight-space averaging as a less expensive
alternative to prediction-space ensembling. Together, these studies establish that ensemble
diversity matters empirically while also showing that nominal diversity in seeds or models
does not guarantee diversity in the errors reached by optimization.

The negative disagreement-selection result is also not unprecedented.
\citet{gleave2022uncertainty} train bootstrap ensembles of language reward models and find
that active learning driven by ensemble uncertainty does not outperform random sampling.
Their estimated epistemic uncertainty is only weakly related to model error, which they
attribute in part to the similarity of members fine-tuned from one underlying language
model. Our DPO pilot differs in both intervention and outcome: it selects model-generated
preference data and measures transfer after policy optimization. Nevertheless, its
three-seed result should be interpreted as a scoped instance of the same broader warning,
not as the first negative result for disagreement-based acquisition.

These works motivate rather than invalidate our analysis. We do not propose that an ensemble
mean is optimal, and we do not claim that covariance is a complete model of adversarial
optimization. We identify the projected response-dependent error governing a fixed
aggregation rule under finite search, derive selected-response and regret guarantees, and
state the information that must come from outside the ensemble before its common-mode risk
can be estimated.

\paragraph{Language-model judges and correlated errors}

RewardBench evaluates reward models on challenging chosen--rejected triples spanning chat,
reasoning, and safety, whereas JudgeBench emphasizes difficult response pairs for which
preference and objective correctness can diverge
\citep{lambert2025rewardbench,tan2024judgebench}. These benchmarks reveal distinct evaluator
failure modes, but static accuracy on either benchmark is not a substitute for evaluating
responses generated by a policy that has been optimized against the evaluator.

Several studies explain why a held-out judge should not automatically be
treated as ground truth. \citet{panickssery2024llm} document self-preference: evaluators can
recognize and favor outputs from related models even when human raters judge the outputs to
be equally good. \citet{dorner2025limits} establish a complementary limitation on scalable
evaluation. When the judge is no more capable than the evaluated model, a small amount of
ground-truth data cannot in general produce an arbitrarily large reduction in evaluation
sample complexity through debiasing alone.

Correlated wrong answers make the same concern visible at the panel level. Across more than
350 language models, \citet{kim2025correlated} find substantial dependence in model errors;
architectural or provider diversity does not guarantee independent failures.
\citet{goel2025great} likewise connect similarity between models to correlated oversight
failures and judge preferences. These findings motivate evaluating error covariance on
responses reachable by the optimized generator, rather than inferring independence from
model names, providers, or checkpoint counts.

Most directly, \citet{kohli2026nine} apply the Kish design effect to a panel of nine frontier
judges from seven model families and find only about two effective independent votes. Their
analysis concerns discrete panel reliability and the gap from an independent Condorcet model.
Equation~\eqref{eq:effective-j} uses the same design-effect algebra for continuous residual
scores. Our incremental step is to place the residual retained by aggregation inside an
optimization-induced selection event and bound both the selected response's overstatement
and its quality regret.

Concurrent work by \citet{chen2026combining} studies a different but adjacent failure object:
the probability that every member of a model pool gives the wrong discrete answer. It proves
that pairwise correlation cannot identify this all-wrong tail and validates the distinction
across a large model pool. That result reinforces a boundary already explicit here: covariance
alone is not a tail guarantee. Our object is the continuous residual projected through a reward
aggregator, and our finite-search bounds require a joint moment-generating-function condition;
the two analyses should not be collapsed into one correlation-only claim.

CARE takes a different next step. \citet{zhao2026care} model static judge scores through
latent quality and shared confounders and prove identifiability and finite-sample recovery
without gold labels under their structural assumptions. We do not offer a competing
latent-variable estimator. Proposition~\ref{prop:nonidentifiability} instead states the
no-assumption boundary: an unrestricted response-dependent common mode cannot be identified
from internal judge scores alone. Proposition~\ref{prop:proxy-contamination} then shows how
one common audit practice biases the relevant covariance, while
Corollary~\ref{cor:two-anchor} gives a correction when two anchor errors satisfy explicit
independence conditions. Structured latent models and external-anchor constructions are
therefore complementary ways to cross the same identification boundary.

\paragraph{Disagreement and active learning}

Query-by-committee selects examples on which members of a committee disagree
\citep{seung1992query}. BALD gives a Bayesian interpretation in which acquisition reflects
information about model parameters \citep{houlsby2011bayesian,gal2017deep}. Panel-based
language-model evaluation applies a related intuition: \citet{verga2024replacing} show that
a jury of smaller, diverse-family evaluators can outperform a single large judge on several
benchmarks. In each case, disagreement is useful when it is coupled to the error or
uncertainty that the intervention is meant to reduce.

The reward-hacking setting adds a specific caution. A committee can disagree because its
members make different errors, and that idiosyncratic variation is exactly what averaging
can suppress. Yet all members may also respond to the same spurious feature. Such an error
changes the ensemble mean without generating internal disagreement. In the projector
notation of Proposition~\ref{prop:disagreement}, the shared component lies in the
aggregation direction and is annihilated by $P$. Low disagreement is consequently evidence
of internal consistency, not a certificate of external correctness.

This distinction does not make disagreement useless. It can identify ambiguous prompts,
poorly calibrated score regions, examples on which additional labels are likely to change
the committee, or cases dominated by member-specific noise. What it cannot do by itself is
rank common-mode errors that all judges endorse. The value of disagreement-based selection
therefore depends on the downstream target: it may improve calibration or label efficiency
without reducing the error exploited by an ensemble mean.

\paragraph{Selective prediction and disagreement filtering}

Selective prediction allows a model to abstain on low-confidence cases, trading coverage for
risk \citep{geifman2017selective}. Disagreement filtering can be viewed as a form of
selective evaluation in which the system refuses to trust examples on which its judges are
inconsistent. Standard risk--coverage reasoning requires the confidence score to rank the
error of interest at least approximately: abstaining must remove errors faster than it
removes correct predictions.

A common-mode evaluator failure can violate this premise. Every judge may be confident and
mutually consistent on the same wrong response. Filtering high-disagreement examples can
then remove genuinely difficult, idiosyncratic cases while retaining the confidently wrong
cases that matter most under optimization. The issue is not merely a different point on a
risk--coverage curve; the internal abstention score may be insensitive to the relevant risk
component. A valid selective-evaluation system therefore needs an external error signal or
a structural model linking disagreement to common-mode risk.

\paragraph{Preference optimization without explicit reward models}

DPO reparameterizes preference optimization so that a policy can be trained directly from a
preference dataset without an explicit online reward-model stage
\citep{rafailov2023direct}. Removing that stage does not remove dependence on the quality of
the preference signal. If model-generated labels contain a correlated bias, the bias can be
propagated into the policy even when no scalar reward model is optimized online. The
overoptimization results of \citet{rafailov2024scaling} make this connection empirical.

We use DPO as a controlled setting in which the acquisition rule can change while the pair
budget and optimization procedure remain fixed. The pilot asks whether selecting prompts
with greater calibrated disagreement changes transfer to an external judge. It does not
claim that the DPO objective itself follows from the covariance theorem, nor that a negative
pilot result generalizes to other preference optimizers, larger models, or acquisition
budgets.

\paragraph{Positioning and research gap}

The literature establishes a progression that constrains our novelty claim. Proxy rewards
can be overoptimized \citep{gao2023scaling,rafailov2024scaling}; reward-model ensembles can
mitigate but not eliminate this behavior
\citep{coste2024ensembles,eisenstein2023helping,rame2024warm}; disagreement is a classical
uncertainty and acquisition signal
\citep{seung1992query,houlsby2011bayesian,gal2017deep}; and recent work directly measures or
models correlated judge errors \citep{kim2025correlated,goel2025great,kohli2026nine,zhao2026care}.
We therefore do not claim to introduce any of those ideas.

The contribution is the connection between their remaining gaps in a finite-search setting.
The projector decomposition separates the error changed by aggregation from the error
observed as disagreement while retaining judge-specific prompt offsets. The identification
result shows why internal consensus alone cannot recover the common mode. The search theorem
turns the retained direction into guarantees for the response actually selected and for its
quality regret, rather than only for the maximum error among searched candidates. Finally,
the anchor analysis distinguishes true common covariance from rank-one contamination induced
by measuring all judges against the same noisy proxy. These results provide an
optimization-aware audit framework; they do not replace conservative aggregation,
latent-confounder modeling, or external evaluation.

\begin{table}[t]
\centering
\small
\setlength{\tabcolsep}{3.2pt}
\resizebox{\columnwidth}{!}{%
\begin{tabular}{llccccc}
\toprule
Generator & Panel & $v$ & $\widehat J_{\rm eff}^{\rm het}$ & sel.\ over-rwd.\ @$K{=}32$ & regret & bench.\ frac. \\
\midrule
SmolLM2 & $\{D_1\}$ & 1.80 & 1.00 & 1.19 [1.02,1.37] & 1.80 & 0.95 \\
SmolLM2 & $\{D_1,D_3\}$ & 1.49 & 1.16 & 1.00 [0.82,1.19] & 1.72 & 0.94 \\
SmolLM2 & $\{D_1,D_2\}$ & 1.55 & 1.10 & 1.03 [0.85,1.21] & 1.64 & 0.93 \\
SmolLM2 & $\{D_1,D_3,G,M\}$ & 1.09 & 1.56 & 0.80 [0.64,0.95] & 1.62 & 0.94 \\
\midrule
Qwen-0.5B & $\{D_1\}$ & 2.07 & 1.00 & 1.27 [1.08,1.46] & 1.72 & 0.96 \\
Qwen-0.5B & $\{D_1,D_3\}$ & 1.72 & 1.13 & 0.87 [0.70,1.02] & 1.44 & 0.97 \\
Qwen-0.5B & $\{D_1,D_2\}$ & 1.66 & 1.09 & 1.06 [0.88,1.22] & 1.63 & 0.96 \\
Qwen-0.5B & $\{D_1,D_3,G,M\}$ & 1.28 & 1.49 & 0.69 [0.55,0.83] & 1.45 & 0.98 \\
\bottomrule
\end{tabular}
}
\caption{Multi-family audit on the locked test split ($K{=}32$ cells shown;
brackets are marginal prompt-bootstrap $95\%$ CIs for selected-response
overstatement). Here $\widehat v$ is a pooled projected residual variance and
$\widehat J_{\rm eff}^{\rm het}$ is the heterogeneous variance-equivalent
panel size defined in Equation~\eqref{eq:heterogeneous-effective-j}.
``bench.\ frac.'' is the descriptive fraction of prompts whose realized error
envelope is below $\sqrt{2\widehat v\log K}$. It has no nominal coverage
interpretation and is not a test of Equation~\eqref{eq:high-prob-bound}.
Panels name the judges actually evaluated: $D_1$--$D_3$ are DeBERTa
checkpoints, $G$ is Gemma-2B, $M$ is Mistral-7B; both $J{=}2$ rows are
within-DeBERTa pairs.}
\label{tab:multifamily}
\end{table}

\section{Archived Multi-Family Audit in Full}
\label{app:multifamily-full}

A preliminary two-judge envelope audit (now superseded; Appendix~\ref{app:bound-table}) found the realized proxy-relative envelope below the Gaussian plug-in curve in every $(J,K)$ cell. We subsequently scaled the evaluation to five eligible models from three families ($D_1$--$D_3$ OpenAssistant DeBERTa, Gemma-2B RM $G$, and Mistral-7B RM $M$), together with a held-out Llama-family proxy excluded from every ensemble. The evaluation used two generators, $J\in\{1,2,4\}$, and $K\in\{2,\dots,32\}$; covariance was estimated on a calibration split, whereas all reported quantities were computed on a locked 200-prompt test split (Table~\ref{tab:multifamily}; artifact manifest in Appendix~\ref{app:artifact-manifest}). The archived run evaluated one panel per row---$\{D_1\}$, the within-DeBERTa pairs, and $\{D_1,D_3,G,M\}$---and therefore does not support same- versus cross-family comparisons; the earlier ``mixed'' finding is accordingly withdrawn. The all-subset audit of \S\ref{sec:all-subset-audit} addresses this limitation using prompt-level records.

Three observations nevertheless remain. First, the projected scale decreases with panel size ($v$: $1.80\!\to\!1.09$ and $2.07\!\to\!1.28$ from $J{=}1$ to $J{=}4$), while $\widehat J_{\rm eff}^{\rm het}\approx1.5$ at $J{=}4$ (Equation~\eqref{eq:heterogeneous-effective-j}), indicating that correlation forfeits most of the nominal averaging benefit. Second, selected-response overstatement at $K{=}32$ decreases with $J$ ($1.19\!\to\!0.80$, $1.27\!\to\!0.69$), whereas target regret improves more modestly ($1.80\!\to\!1.62$, $1.72\!\to\!1.45$). This pattern is consistent with the theoretical separation between suppressing proxy over-scoring and improving selection. Because the archived summary lacks paired prompt-level contrasts, these decreases are descriptive rather than inferential; the records audit above provides the paired analysis. Third, the fraction of realized envelopes below the expectation-scale benchmark $\sqrt{2\widehat v\log K}$ ranges from $0.81$ to $0.99$. This quantity is a pass fraction rather than a coverage probability: it has no confidence parameter and uses an uncertified $\widehat v$. Consequently, it neither demonstrates undercoverage nor validates Theorem~\ref{thm:search-bound}, and we make no nominal coverage claim for the real-model experiments.

Finally, we do not report a common-mode comparison across $J$. The archived rates were computed using within-test-cell thresholds rather than calibration-frozen thresholds, and disagreement is identically zero at $J{=}1$ (yielding a rate of $0.25$ by construction). A frozen-threshold recomputation requires per-prompt records that are unavailable in the archived summary. Artifact layout, contents, and remaining gaps are documented in Appendix~\ref{app:artifact-manifest}.

\section{Anchor Sensitivity and Verifier-Backed Evaluation in Full}
\label{app:anchor-sensitivity}

The audits above measure error against a held-out learned judge, exactly the
practice Proposition~\ref{prop:proxy-contamination} warns about. We therefore
compare two held-out learned anchors from different families (Llama-3-8B RM,
Mistral-7B RM; fixed Qwen2.5-1.5B generator, $K{=}8$) with objective
verification where the task admits it---a sensitivity analysis, not an
instantiation of Corollary~\ref{cor:two-anchor}, since family difference
does not imply conditional independence. On GSM8K (exact-match anchor,
prevalence $0.37$) the anchors correlate at Spearman $0.42$ and rank a
verified-correct candidate first on $0.73$/$0.66$ of solvable prompts
(DeBERTa ensemble: $0.62$). On HumanEval the verifiable anchor is
\emph{degenerate} (zero passing suites), reported as such. On open-ended
prompts the anchors correlate \emph{negatively} (Spearman $-0.21$) and mean
selected-response regret depends strongly on the anchor ($3.8$ vs $6.1$ on
GSM8K; $4.0$ vs $7.3$ open-ended). Only GSM8K supplies a non-degenerate
verifier-backed target; the negative correlations reinforce the anchor-validity
concern, so the two-anchor identity and its UCB are verified empirically
only in the controlled synthetic experiment.

\section{Two-Judge Error-Envelope Audit Table}
\label{app:bound-table}

Setup (moved from \S\ref{sec:bound-validation}): candidates from
SmolLM2-360M-Instruct, two $z$-calibrated reward-model judges, a stronger
held-out proxy $\widetilde q$ excluded from the ensemble, and the mean over
$120$ prompts of the proxy-relative envelope $\max_{k\leq K}\bar e(x,a_k)$---an
upper bound on the selected response's error, not the selected-response
statistic itself.  With plug-in variances $\widehat v_{J=1}=0.708$ and
$\widehat v_{J=2}=0.696$, the observed envelope stays below the Gaussian
plug-in curve in every $(J,K)$ cell.  The magnitudes are not precisely
predicted---the plug-in curves are loose and the run kept no prompt-level
intervals---and an equal-variance fit ($\sigma^2\approx0.84$,
$\rho\approx0.65$) predicts disagreement $0.147$ vs.\ the measured $0.176$:
qualitatively consistent, not exact, with
Proposition~\ref{prop:proxy-contamination} explaining why the shared proxy
prevents reading the fitted correlation as ground-truth common-mode error.

\begin{table}[h]
\centering
\small
\begin{tabular}{ccccc}
\toprule
$J$ & $K$ & observed envelope & plug-in & obs./plug-in \\
\midrule
$1$ & $2$  & $0.233$ & $0.991$ & $0.235$ \\
$1$ & $4$  & $0.491$ & $1.401$ & $0.350$ \\
$1$ & $8$  & $0.794$ & $1.716$ & $0.463$ \\
$1$ & $16$ & $1.002$ & $1.981$ & $0.506$ \\
$1$ & $32$ & $1.164$ & $2.215$ & $0.525$ \\
\midrule
$2$ & $2$  & $0.172$ & $0.982$ & $0.175$ \\
$2$ & $4$  & $0.440$ & $1.389$ & $0.317$ \\
$2$ & $8$  & $0.730$ & $1.701$ & $0.429$ \\
$2$ & $16$ & $0.898$ & $1.965$ & $0.457$ \\
$2$ & $32$ & $1.018$ & $2.196$ & $0.464$ \\
\bottomrule
\end{tabular}
\caption{Finite-search error-envelope audit across all reported $(J,K)$ cells. Observed proxy-relative envelopes remain below the Gaussian plug-in curves $\sqrt{2\widehat v\log K}$ computed from $\widehat v_{J=1}=0.708$ and $\widehat v_{J=2}=0.696$. The table supports numerical consistency with the upper envelope; it does not show equality, identify a $\sqrt{\log K}$ scaling law, or isolate the causal effect of correlation from judge identity.}
\label{tab:bound-validation}
\end{table}

\section{Synthetic Stress-Test Details}
\label{app:synthetic-stress}

\begin{figure*}[t]
    \centering
    \includegraphics[width=0.96\textwidth]{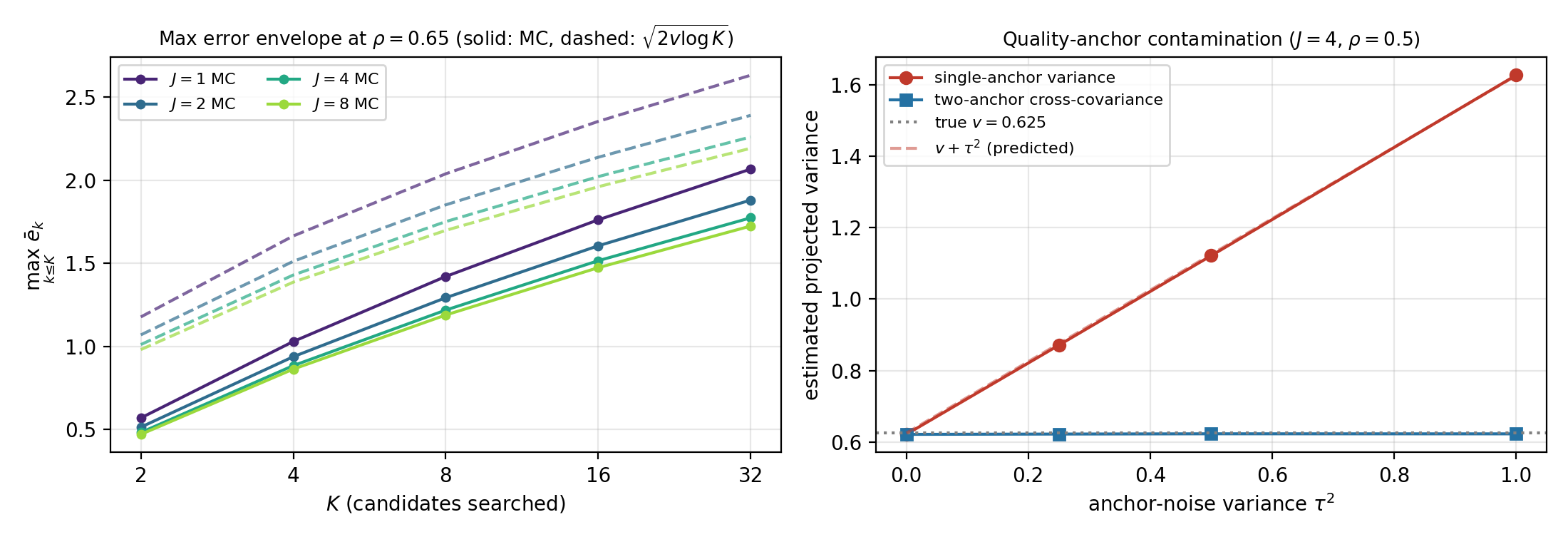}
    \caption{Fixed-seed synthetic stress test. Left: for $\rho=0.65$, solid curves are Monte Carlo means of $\max_{k\leq K}\bar e_k$ and dashed curves are $\sqrt{2v\log K}$. Right: one noisy anchor inflates projected variance by approximately $\tau^2$, while the cross-covariance from independent anchors remains near the true value. Because the data are generated from the assumed Gaussian model, this is an implementation and assumption stress test rather than real-model validation.}
    \label{fig:synthetic-stress}
\end{figure*}

The synthetic audit is fully specified by Equation~\eqref{eq:synthetic-model}. For each of the
$24$ $(J,\rho)$ settings, one draw contains $50{,}000$ prompts and $32$ candidates per prompt.
The $K\in\{2,4,8,16,32\}$ conditions use nested prefixes of the same candidate tensor. The
target qualities and all common and judge-specific error variables are independent standard
Gaussians. Ensemble selection maximizes $q_k+\bar e_k$; the reported quantities are
$\max_k\bar e_k$, $\bar e_{k_R}$, and $\max_kq_k-q_{k_R}$. Their reference curves are the
corresponding terms in Theorem~\ref{thm:search-bound} with
$v=\rho+(1-\rho)/J$.

The anchor experiment uses $200{,}000$ independent scalar projected errors with $J=4$ and
$\rho=0.5$. Two independent Gaussian anchor errors are added at each value of
$\tau^2\in\{0,.25,.50,1.00\}$. The single-anchor estimator is the sample variance of one
proxy-relative projected residual; the corrected estimator is the sample cross-covariance
between the two proxy-relative residuals.  With true projected variance $0.625$, the
single-anchor estimate rises from $0.621$ to $0.871$, $1.121$, and $1.626$ as $\tau^2$
increases, while the two-anchor estimates remain $0.621$, $0.622$, $0.623$, and $0.623$. The experiment uses no fitted parameters and no
discarded runs. The accompanying simulation script and two CSV outputs reproduce
Figure~\ref{fig:synthetic-stress} from seed 20260802.

This stress test is narrower than a model evaluation. Gaussian draws make the
sub-Gaussian proxy exact, while the construction makes the anchor errors independent.
The stress test can reveal implementation mistakes, arithmetic inconsistencies, or a mismatch
between a theorem and its claimed statistic. It cannot establish that learned judges satisfy
the assumptions, which is why the real-model audit and the missing confirmatory experiments
remain separately identified.

\section{Exploratory DPO Pilot}
\label{sec:selection-result}

This appendix retains the original controlled DPO experiment, while narrowing its conclusion to the available evidence. Two training arms start from the same Qwen2.5-1.5B-Instruct checkpoint and use the same DPO budget over three rounds. One arm selects prompts with high calibrated cross-judge disagreement; the other selects a size-matched random sample. Preference labels in both arms come from the calibrated ensemble mean. Evaluation uses identical held-out prompts and a held-out judge. The full acquisition protocol and the additional matching controls originally specified for a confirmatory study appear in Appendix~\ref{app:dpo-protocol}.

Across three paired training seeds and 120 evaluation prompts, the held-out-score differences are $0.035$, $0.033$, and $0.002$. Their mean is $0.023$ with standard error $0.011$. The exact sign-flip test gives one-sided $p=0.125$ and two-sided $p=0.250$; the previously reported $p=0.13$ was the rounded one-sided value. A $t$ interval with two degrees of freedom is $[-0.023,0.069]$. The exploratory three-seed DPO pilot is summarized in Table~\ref{tab:dpo-pilot}.

\begin{table}[t]
\centering
\small
\begin{tabular}{lc}
\toprule
Experimental unit & disagreement $-$ random \\
\midrule
Seed 1 & $+0.035$ \\
Seed 2 & $+0.033$ \\
Seed 3 & $+0.002$ \\
\midrule
Mean $\pm$ SE & $+0.023\pm0.011$ \\
95\% $t$ interval & $[-0.023,0.069]$ \\
Exact two-sided $p$ & $0.250$ \\
\bottomrule
\end{tabular}
\caption{Seed-level DPO pilot. All three estimates are positive, but the interval is too wide to establish either a reliable improvement or practical equivalence. Prompts are repeated measurements within a trained model and are not treated as independent experimental units.}
\label{tab:dpo-pilot}
\end{table}

The pilot yields no statistically reliable evidence that high-disagreement selection improves transfer at this scale. It also does not show that the intervention has no practically meaningful effect. The pooled-prompt analysis reported $p=0.41$, but it is secondary because prompts sharing a trained model are not independent experimental units. A confirmatory negative result would require a predeclared smallest effect of interest, an equivalence test, and a seed count chosen by power analysis. Appendix~\ref{app:selection-analysis} develops the possible interpretations without treating planned controls as completed results. The accompanying seed-summary artifact records the three values and the exact analysis; candidate-level and training logs were not present in the supplied artifact and remain necessary for full reproduction.

\section{Disagreement-Selection Protocol}
\label{app:dpo-protocol}

The intervention is calibrated disagreement selection. At each round, the current generator
samples at least two responses for every prompt under a fixed decoding configuration. Each
training judge scores the same responses, and all transformations that place those scores on
a common scale are fitted on a calibration split disjoint from prompt selection, policy
training, and final evaluation. Reusing the acquisition or evaluation prompts for calibration
would allow the score mapping itself to absorb part of the intervention.

For response $a_k$ to prompt $x$, the algorithm computes the calibrated disagreement in
Equation~\eqref{eq:disagreement}. The prompt-level acquisition score is the average of this
quantity across its $K_r$ sampled responses. Averaging avoids making the acquisition decision
depend on one anomalous sample, while keeping the response-level scores available for
diagnostic analysis. The disagreement arm selects prompts from a prespecified top quantile.
The random arm draws a size-matched subset from the same eligible pool, and the matched
controls draw from that pool subject to the constraints in Table~\ref{tab:controls}.

Preference construction is held fixed after acquisition. The calibrated ensemble mean labels
each response pair, a prespecified margin removes ties and near-ties, and resampling restores
the same number of retained pairs in every arm. Thus the intended intervention is the
distribution of prompts entering preference construction, not the number of labels, the
confidence of those labels, or the amount of optimization. In later rounds, candidates are
generated by the current policy rather than the initial checkpoint, so the acquisition rule
is evaluated under the distribution shift it helps create.

\begin{algorithm}[t]
\caption{Calibrated disagreement-selection DPO arm}
\label{alg:selection}
\begin{algorithmic}[1]
\REQUIRE initial policy $\pi_{\theta_0}$; prompt pool $\calX_{\mathrm{train}}$
\REQUIRE calibrated judges $1,\ldots,J$; rounds $T$; candidates per prompt $K_r$
\REQUIRE pairs per round $M$; selection rule $S$
\STATE set $\theta\leftarrow\theta_0$
\FOR{$t=1,\ldots,T$}
\STATE sample $K_r$ responses per prompt using the fixed decoding configuration
\STATE compute calibrated judge scores and per-prompt disagreement $\widehat D_t(x)$
\STATE select exactly $M$ preference pairs according to $S$ and its matching constraints
\STATE label each pair by the calibrated ensemble mean
\STATE discard pairs below the tie margin and resample to retain $M$ pairs
\STATE update $\theta$ with the fixed DPO objective, reference model, and optimizer schedule
\ENDFOR
\RETURN final policy $\pi_\theta$
\end{algorithmic}
\end{algorithm}

The random comparison is the minimum causal control. It measures the effect of performing the
same amount of preference optimization without targeting disagreement. Pair counts, generated
tokens, judge calls, DPO steps, decoding settings, and tie-handling rules must be identical.
Otherwise, an apparent acquisition gain could be explained by a larger data or compute budget.

Length matching addresses a more specific alternative. High-disagreement prompts or responses
may be longer, more likely to reach truncation limits, or more likely to elicit verbose outputs.
Matching prompt and response-length distributions tests whether the result is caused by those
properties rather than evaluator uncertainty. Initial-reward matching distinguishes
disagreement from ordinary hard-example mining: the control should reproduce the initial
ensemble-score distribution without conditioning on cross-judge spread.

Update-norm matching addresses the possibility that selected examples merely induce larger
optimization steps. It matches either gradient norms during training or the resulting
parameter displacement under a declared convention. Finally, comparing mean aggregation with
a minimum or another conservative rule asks whether any benefit is specific to the mean rather
than to the selected data. These controls isolate different mechanisms and are not
interchangeable.

Only the random comparison is present in the supplied pilot outputs. The length-, reward-, and
update-norm-matched arms and the conservative-aggregation comparison therefore remain
requirements for a confirmatory study. They are retained here as part of the full experimental
design, not presented as completed evidence.

\section{Experimental Design and Reporting Requirements}
\label{app:experimental-design}
\label{sec:design}

The intended unit of randomization is the training seed. Each seed initializes all arms from
the same checkpoint and uses identical train, calibration, and evaluation splits. Randomness
in candidate sampling, pair construction, data ordering, and optimization should be paired
where doing so does not leak the acquisition decision. Across arms, the number of preference
pairs, DPO steps, generated tokens, and judge queries must be matched unless a separate
cost-normalized comparison is explicitly reported.

This distinction determines the statistical analysis. Evaluation prompts are repeated
measurements within a trained policy, not independent replications of training. Seed-level
paired differences are therefore the primary experimental units. Prompt bootstrap or
permutation intervals can describe uncertainty conditional on the trained policies, but they
cannot replace inference across independently trained seeds. Hyperparameters and stopping
rules must be selected without examining the final paired outcomes. For reproducibility, the minimum information that must be reported for the generator, decoding procedure, judges, DPO configuration, selection procedure, evaluation protocol, and statistical analysis is summarized in Table~\ref{tab:repro-spec}. The checklist specifies the required fields, while the archival configuration should provide the actual values for each field.

\begin{table*}[t]
\centering
\small
\begin{tabular}{lll}
\toprule
Component & Required specification & Reason \\
\midrule
Generator & checkpoint hash, tokenizer, chat template & prevents model ambiguity \\
Decoding & temperature, top-$p$, max tokens, stop strings & controls the candidate distribution \\
Judges & checkpoint hashes, calibration split, score transform & makes aggregation interpretable \\
DPO & reference model, $\beta$, optimizer, LR, batch size, epochs & makes training reproducible \\
Selection & acquisition statistic, quantile, tie margin, resampling & isolates the intervention \\
Evaluation & independent anchors, verifiable metrics, prompt split & limits circular validation \\
Statistics & seed-level test, interval, equivalence margin, correction & avoids pseudo-replication \\
\bottomrule
\end{tabular}
\caption{Minimum information required for a reproducible implementation. A checklist is not a substitute for reporting the actual values; every entry must be instantiated in the archival configuration.}
\label{tab:repro-spec}
\end{table*}

The primary outcome should be selected-response target quality, accompanied by the training
ensemble score and the proxy--anchor gap. Reporting all three distinguishes an intervention
that genuinely improves external quality from one that merely increases the score of the
evaluators used to generate its labels. When candidates are retained, the analysis should
also report regret relative to the best candidate under the external anchor, matching the
quantity in Theorem~\ref{thm:search-bound}.

A held-out judge is itself imperfect and should not be described as ground truth. Verifiable
mathematical answers, unit-tested executable code, exact structured-output checks, or other
task-specific validators provide scalable anchors when human evaluation is unavailable.
RewardBench-style preference triples and JudgeBench-style correctness pairs can complement
these outcomes because they probe different evaluator failures
\citep{lambert2025rewardbench,tan2024judgebench}. Static benchmark accuracy nevertheless
cannot replace evaluation of the optimized policy's own generations, where the distribution
shift induced by search or DPO is present.

\begin{table*}[t]
\centering
\small
\begin{tabular}{L{0.23\linewidth}L{0.17\linewidth}L{0.25\linewidth}L{0.23\linewidth}}
\toprule
Comparison & Variable isolated & Must be matched & Interpretation if tied \\
\midrule
Disagreement vs. random & acquisition rule & pairs, tokens, judge calls & no detected acquisition benefit \\
Disagreement vs. length-matched & verbosity or truncation & prompt and response length & apparent gain may be length-driven \\
Disagreement vs. reward-matched & initial difficulty & initial ensemble score & apparent gain may be hard-example mining \\
Disagreement vs. norm-matched & update magnitude & gradient or parameter norm & apparent gain may be step-size driven \\
Mean vs. minimum ensemble & aggregation rule & selected data and budget & tests a conservative reward baseline \\
\bottomrule
\end{tabular}
\caption{Prespecified controls for attributing an improvement to calibrated disagreement selection. The current pilot executes only the first row; the table defines the complete confirmatory design and does not report additional results.}
\label{tab:controls}
\end{table*}

\subsection{Evaluation across judge and generator families}

The two-judge pilot cannot separate the effect of adding a member from the identity of the
member added. A confirmatory matrix should therefore vary ensemble size and the source of
diversity. At minimum, it should distinguish members that share a pretrained checkpoint but
use different reward-model fine-tuning seeds, members with different pretraining seeds, and
members from different architectures or providers. The relevant comparison is measured
error covariance on the policy's candidate distribution, not the nominal category alone.

The generator side should vary both model family and search pressure. For best-of-$K$ audits,
the same base candidates should be reused across nested budgets whenever possible, so changes
with $K$ are not confounded by independent sampling noise. For trained policies, evaluation
should include the initial generator and every final arm under identical decoding. Reporting
$J\in\{1,2,4,\ldots\}$ and several $K$ values permits direct comparison of nominal ensemble
size, effective size, and the search budget that acts on the residual projected error.

Every judge subset must be calibrated without access to its test responses. Covariance,
disagreement, external-anchor error, and judge-call cost should be reported for each subset.
This design can reveal whether cross-family diversity genuinely reduces the aggregation
direction or merely increases orthogonal disagreement. It also prevents one favorable pair
of judges from being treated as evidence for a general ensemble-size law.

\subsection{Diagnostics connecting acquisition to error}

The first diagnostic plots calibrated disagreement against error under an independent
anchor. In addition to a correlation coefficient, it should display the full joint
distribution and the low-disagreement, high-error quadrant that exposes common-mode failures.
Stratification by prompt length, response length, task, and initial ensemble score tests
whether the acquisition statistic is acting as a proxy for a simpler observable property.

The second diagnostic compares the projected covariance in
Equation~\eqref{eq:matrix-var} with selected-response overstatement across judge subsets and
search budgets. The unit of analysis and uncertainty interval must be declared: prompt-level
intervals should be clustered by prompt, while comparisons between trained policies must
retain seed-level replication. A third calibration diagnostic should compare raw and
calibrated score distributions, because a disagreement trend that disappears after placing
judges on a common scale is a units artifact rather than evidence of epistemic diversity.

All diagnostics should report judge calls, generated tokens, and wall-clock or accelerator
cost. A method that reduces error only by multiplying evaluator cost may still be useful, but
its advantage is different from a covariance-efficient ensemble. Diagnostic plots are
mechanism checks; they do not replace the primary external-quality comparison.

\section{Interpreting the DPO Pilot}
\label{app:selection-analysis}

The three seed-level differences are positive, but their interpretation is constrained by
the size of the experiment. The mean difference of $0.023$ is a point estimate, not evidence
that the acquisition rule reliably improves transfer. The $t$ interval crosses zero and
includes effects that would be practically meaningful in either direction, while the exact
two-sided randomization test has only four attainable levels below one. Conversely, failure
to reject zero is not evidence that the methods are equivalent. An equivalence claim would
require a prespecified smallest effect of interest and an interval lying entirely inside the
corresponding equivalence region.

The result is compatible with the covariance analysis but is not predicted by it. The theory
shows that disagreement observes the component orthogonal to mean aggregation. It does not
say that examples with large orthogonal error contain no useful training signal. Such examples
could still improve calibration, expose ambiguous preferences, diversify response styles, or
produce harder preference pairs. Whether those benefits transfer to an external evaluator is
an empirical question determined by the data distribution, label mechanism, model capacity,
and optimization budget.

Several alternative mechanisms remain unresolved in the current pilot. High-disagreement
prompts may be longer or harder, may receive lower initial ensemble scores, or may generate
larger DPO updates. A held-out reward model can also share errors with the training judges,
which would make transfer under that judge an incomplete measure of common-mode failure. The
matched controls and independent anchors specified above are needed to distinguish these
possibilities.

The defensible conclusion is therefore deliberately local: under the reported checkpoint,
three-round DPO budget, acquisition procedure, and three paired seeds, the experiment does
not provide statistically reliable evidence that disagreement selection improves held-out
judge score. It neither establishes a benefit nor rules out a practically important one, and
it does not support a universal claim about disagreement-based data selection.

\section{Claim Boundaries and Failures}
\label{app:claim-boundaries}

The formal claim is deliberately conditional. Under calibrated scores, a declared candidate
distribution, and a joint sub-Gaussian error proxy, the error projected onto a fixed
aggregation direction controls selected-response overstatement and target-quality regret.
Covariance alone does not imply the required tail condition. Heavy-tailed or adversarially
dependent errors may make a covariance plug-in severely optimistic even when the second
moment is estimated accurately. The empirical curves in Table~\ref{tab:bound-validation}
must therefore remain working-model diagnostics unless the proxy matrix or tail condition is
independently justified.

Finite search is another substantive boundary. The union-bound argument allows dependence
among the $K$ candidate errors. Corollary~\ref{cor:adaptive-search} also allows proposal
distributions to react predictably to earlier evaluator outputs, but only when the projected
error remains conditionally centered and sub-Gaussian after conditioning on that history.
Repeated policy training can invalidate precisely this condition by moving toward regions in
which calibration fails. The corollary is therefore a finite adaptive-search result, not a
certificate for unrestricted policy--evaluator co-adaptation.

The primary structural failure is a response-dependent common error. If every judge rewards
the same spurious feature, the ensemble can agree confidently while assigning the wrong
ordering to candidates. Proposition~\ref{prop:nonidentifiability} shows that the resulting
common mode cannot be recovered from judge scores alone without additional structure. A
prompt-only calibration offset is different: it changes absolute overstatement but cancels
from within-prompt selection. Keeping these notions separate avoids attributing an
optimization failure to a constant that search cannot exploit.

Score calibration introduces a second boundary. Reward models can differ in offset, scale,
nonlinearity, and saturation. Without a common interval interpretation, one judge can
dominate the mean and disagreement can reflect units rather than uncertainty. Global
calibration also does not guarantee that judge-specific prompt offsets vanish, which is why
Proposition~\ref{prop:disagreement} retains the $\norm{P\mathbf b}^2$ term. Calibration must
be audited on held-out data and, when possible, by task and score region rather than only in
aggregate.

The external quality anchor is not exempt from evaluator error. A single held-out reward
model can share training data, architecture, style preferences, or blind spots with the
training ensemble. Even independent anchor noise contributes the rank-one term in
Proposition~\ref{prop:proxy-contamination}; correlated anchor noise can be more difficult to
separate. The two-anchor correction is valid only under its conditional independence
assumptions. Closely related language models should not be called independent anchors merely
because they use different checkpoint names.

Selection itself induces distribution shift. Prompts selected this way may
differ from random prompts in length, difficulty, topic, adversarial structure, initial
reward, or the size of the resulting gradient. If disagreement beats random selection but
not a length- or reward-matched control, the appropriate conclusion is that the observed gain
is explained by that matched property. Table~\ref{tab:controls} is designed to separate these
mechanisms, but only its random comparison was executed in the supplied pilot.

Finally, the analysis does not establish that the ensemble mean is optimal or that
disagreement selection should improve preference optimization. Worst-case,
uncertainty-weighted, covariance-aware, and latent-confounder-aware objectives remain
competing approaches \citep{coste2024ensembles,zhao2026care}. The theoretical contribution
characterizes a chosen aggregation rule and an identification problem; the empirical pilot
tests one acquisition intervention. Neither should be generalized into a universal ranking
of ensemble designs.

\section{Statistical Details for the DPO Pilot}
\label{app:statistics}

For paired seed differences $d=(0.035,0.033,0.002)$, the sample mean is $\bar d=0.02333$, the sample standard deviation is $0.01850$, and the standard error is $0.01068$. Using $t_{0.975,2}=4.303$ gives
\[
\bar d\pm t_{0.975,2}\operatorname{SE}(d)
=[-0.0226,0.0693].
\]
An exact sign-flip randomization test enumerates the eight possible sign assignments. One assignment is at least as large as the observed positive mean, giving one-sided $p=1/8=0.125$; two assignments are at least as extreme in absolute value, giving two-sided $p=2/8=0.250$. With three seeds, exact $p$-values are necessarily coarse. The prompt-pooled $p=0.41$ is retained only as a secondary descriptive analysis because it treats repeated prompts within a trained model as if they supplied independent training replicates.

A confirmatory study should specify a smallest effect of interest before looking at final outcomes. Demonstrating no practically important benefit then requires the confidence interval to fall inside the equivalence region, not merely a failure to reject zero. The pilot interval is too wide for such a conclusion.

\section{Artifact Manifest and Model Inventory}
\label{app:artifact-manifest}

\paragraph{Model inventory}

Table~\ref{tab:model-manifest} lists every model with its role and
aggregation eligibility.  The executed runs loaded checkpoints by repository
name without pinning immutable revision hashes; the hashes therefore cannot
be reconstructed honestly after the fact and are marked as not recorded.  A
rerun that pins revisions is required before hash-level reproducibility can
be claimed.

\begin{table*}[t]
\centering
\small
\resizebox{\textwidth}{!}{%
\begin{tabular}{lllll}
\toprule
ID & Repository & Family & Role & Eligible for aggregation \\
\midrule
$D_1$ & \path{OpenAssistant/reward-model-deberta-v3-large-v2} & DeBERTa & judge & yes \\
$D_2$ & \path{OpenAssistant/reward-model-deberta-v3-large} & DeBERTa & judge & yes \\
$D_3$ & \path{OpenAssistant/reward-model-deberta-v3-base} & DeBERTa & judge & yes \\
$G$ & \path{weqweasdas/RM-Gemma-2B} & Gemma & judge & yes \\
$M$ & \path{weqweasdas/RM-Mistral-7B} & Mistral & judge & yes \\
--- & \path{Ray2333/GRM-Llama3-8B-rewardmodel-ft} & Llama & held-out proxy & no \\
--- & \path{HuggingFaceTB/SmolLM2-360M-Instruct} & SmolLM & generator & --- \\
--- & \path{Qwen/Qwen2.5-0.5B-Instruct} & Qwen & generator & --- \\
--- & \path{Qwen/Qwen2.5-1.5B-Instruct} & Qwen & generator (\S\ref{sec:targetquality}, DPO) & --- \\
\bottomrule
\end{tabular}
}
\caption{Model inventory. Revision hashes were not pinned by the executed
runs and are not recorded; the DeBERTa checkpoints are one family
(three checkpoints), so the eligible pool spans three families
(DeBERTa, Gemma, Mistral). Panels actually evaluated: $\{D_1\}$,
$\{D_1,D_3\}$, $\{D_1,D_2\}$, $\{D_1,D_3,G,M\}$; an all-subsets design
($J{=}2$: three DeBERTa-within-family pairs and all seven cross-family
pairs; $J{=}4$: the three three-family panels) was not executed.}
\label{tab:model-manifest}
\end{table*}

\paragraph {Aggregation baselines (executed)}

For a comparison of aggregation rules on identical candidate tensors,
calibration split, test prompts, and search budgets, the predeclared
baseline set is: each individual judge with any ``best single'' designation
chosen on calibration data only; the uniform mean; minimum aggregation
$\min_j r_j$; uncertainty-penalized aggregation $\bar r-\lambda\sqrt D$ with
$\lambda$ chosen on calibration data only; simplex-constrained covariance
weighting
$\widehat{\mathbf w}\in\arg\min_{\mathbf w\geq0,\one^\top\mathbf w=1}
\mathbf w^\top\widehat\Sigma_{\rm cal}\mathbf w$; and the closest executable
confounder-aware aggregation baseline (CARE-SVD) at its public
implementation revision.  All six are now executed in the all-subset
records audit (\S\ref{sec:all-subset-audit}) with selected target quality,
regret, selected and centered overstatement, disagreement, judge calls, and
prompt-paired intervals against the uniform mean, released per
(panel, $K$, method) cell in the artifact CSVs.

\paragraph{Certification protocol (deferred from \S\ref{sec:real-anchor-certificate}).}

For the one predeclared non-degenerate task (GSM8K; exact match is target
and anchor~1, so $\eta_1\equiv0$ by construction; the held-out GRM-Llama3
reward model is anchor~2), we invoke
Theorem~\ref{thm:bounded-anchor-search} on one independent pair per
certification prompt ($m{=}80$; first two candidates in the frozen order),
all scales fixed before outcomes were inspected ($C_1{=}C_2{=}c{=}2$,
target width $1$; $\delta_{\rm est}{=}\delta_{\rm search}{=}0.025$
pointwise, Bonferroni across the $855$-cell family). Reported exactly as
observed: the certificate is \emph{valid but uninformative in all $855$
cells}. $\widehat v_s$ is small (median $0.017$, max $0.27$) but the
estimation correction
$C_1C_2\sqrt{\log(1/\delta_{\rm est})/2m}\approx0.61$ dominates it, so
every Bernstein and Bennett radius hits its deterministic range cap and
the pass fractions of $1.0$ are trivial, \emph{not} coverage. Reaching
$\widehat v_s$-scale precision at these constants needs
$m\approx10^{5}$ pairs: at paper scale the certificate's contribution is
the honest sizing formula, not a usable bound
(Appendix~\ref{app:anchor-ucb-cells}).

\paragraph{Real-task certificate cells}
\label{app:anchor-ucb-cells}

Table~\ref{tab:anchor-ucb} reports representative certificate cells from
the $855$-cell family (all cells share $m{=}80$, $C_1{=}C_2{=}c{=}2$,
target width $1$; per-cell values, both radius families, and the Bennett
inversion are in \texttt{real\_outputs/anchor\_ucb\_cells.csv}). No cell is
favorable: $\widehat v_s\in[0.009,0.27]$ (median $0.017$) is dominated by
the $m{=}80$ estimation correction, so \emph{every} pointwise and
simultaneous radius---Bernstein and Bennett alike---is capped at its
deterministic range bound in all $855$ cells. The pass fractions of $1.0$
are trivial consequences of the caps, not coverage.

\begin{table}[t]
\centering
\small
\resizebox{\columnwidth}{!}{%
\begin{tabular}{llrrrrrrrrr}
\toprule
Panel & Rule & $K$ & $m$ & $\widehat v_s$ & $v^{\mathrm U}_{\rm pt}$ & $v^{\mathrm U}_{\rm sim}$ & $B_s^{\rm pt}$ & $T_s^{\rm pt}$ & $B_s^{\rm sim}$ & $T_s^{\rm sim}$ \\
\midrule
$J{=}1$ (repr.) & uniform mean & 32 & 80 & 0.0165 & 0.624 & 1.038 & 2.0 & 1.0 & 2.0 & 1.0 \\
$J{=}2$ (repr.) & uniform mean & 32 & 80 & 0.0142 & 0.622 & 1.036 & 2.0 & 1.0 & 2.0 & 1.0 \\
$J{=}3$ (repr.) & uniform mean & 32 & 80 & 0.0159 & 0.623 & 1.038 & 2.0 & 1.0 & 2.0 & 1.0 \\
$J{=}4$ (repr.) & uniform mean & 32 & 80 & 0.0141 & 0.621 & 1.036 & 2.0 & 1.0 & 2.0 & 1.0 \\
$J{=}5$ & uniform mean & 2 & 80 & 0.0096 & 0.617 & 1.031 & 2.0 & 1.0 & 2.0 & 1.0 \\
$J{=}5$ & uniform mean & 8 & 80 & 0.0096 & 0.617 & 1.031 & 2.0 & 1.0 & 2.0 & 1.0 \\
$J{=}5$ & uniform mean & 32 & 80 & 0.0096 & 0.617 & 1.031 & 2.0 & 1.0 & 2.0 & 1.0 \\
\bottomrule
\end{tabular}
}
\caption{Representative anchor-UCB certificate cells ($B_s$: selected-error
radius, capped at $C_1{=}2$; $T_s$: regret radius, capped at the target
width $1$; pt/sim: pointwise $\delta_{\rm est}{=}\delta_{\rm search}{=}0.025$
vs.\ Bonferroni-simultaneous). The estimation correction
$C_1C_2\sqrt{\log(1/\delta_{\rm est})/2m}\approx0.61$ dominates
$\widehat v_s$ at $m{=}80$, so all $855$ cells are range-capped; Bennett
radii equal Bernstein radii at these constants after capping.}
\label{tab:anchor-ucb}
\end{table}

\section{Supplementary Figures}
\label{app:supp}

\begin{figure*}[t]
\centering
\includegraphics[width=0.95\linewidth]{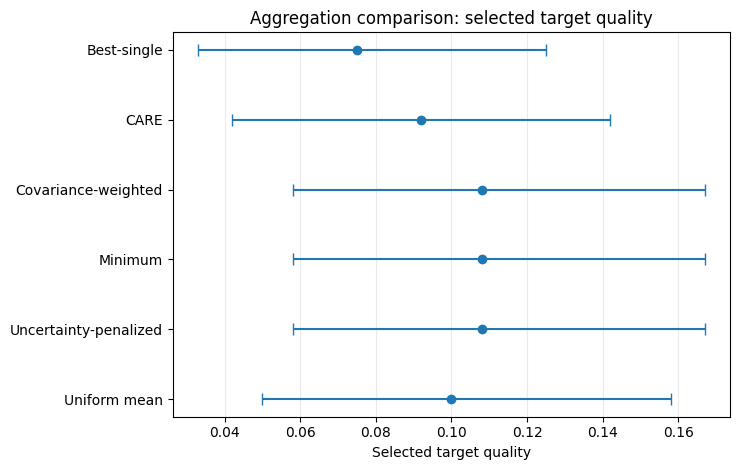}
\caption{Aggregation comparison on selected target quality for the largest eligible panel ($J=5$) and search budget ($K=32$). Points show mean selected target quality and horizontal bars show prompt-paired bootstrap 95\% intervals over the 120 locked test prompts.}
\label{fig:aggregation-quality}
\end{figure*}

Figure~\ref{fig:aggregation-quality} visualizes the corresponding selected-quality estimates and prompt-paired bootstrap intervals.

\begin{figure*}[t]
\centering
\includegraphics[width=0.95\linewidth]{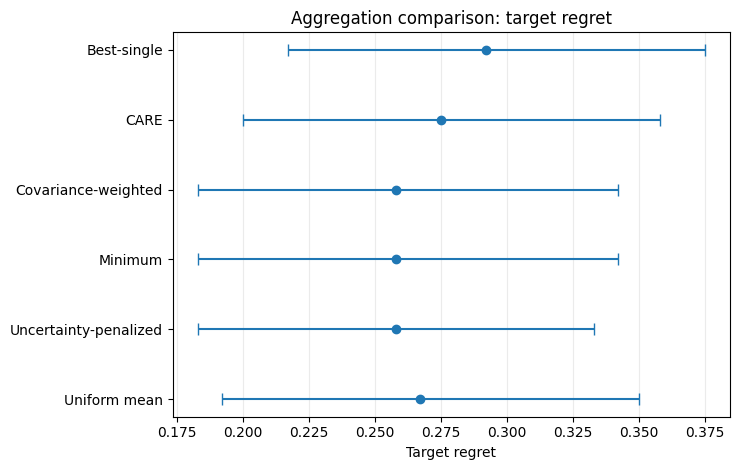}
\caption{Aggregation comparison on target regret for the largest eligible panel ($J=5$) and search budget ($K=32$). Points show mean target regret and horizontal bars show prompt-paired bootstrap 95\% intervals over the 120 locked test prompts.}
\label{fig:aggregation-regret}
\end{figure*}

Figure~\ref{fig:aggregation-regret} shows the corresponding target-regret comparison.

\begin{figure*}[t]
\centering
\includegraphics[width=0.95\linewidth]{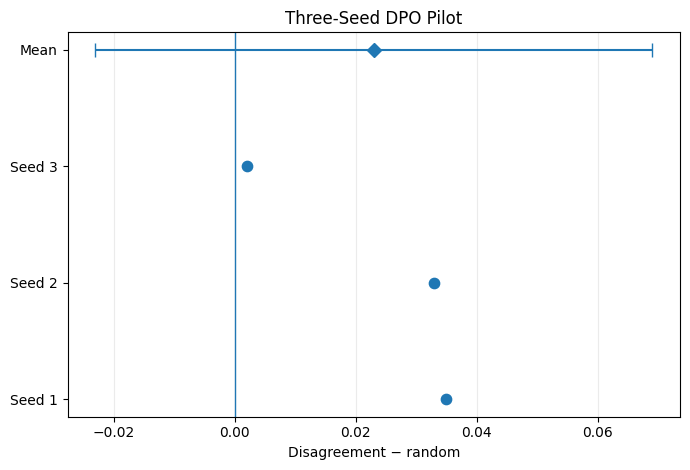}
\caption{Three-seed DPO pilot comparing disagreement selection with random selection. Points show the seed-level differences, and the diamond shows the mean with its 95\% $t$ interval. The interval crosses zero, so the pilot does not establish a reliable improvement or practical equivalence.}
\label{fig:dpo-pilot}
\end{figure*}

Figure~\ref{fig:dpo-pilot} shows the corresponding seed-level differences and their uncertainty interval.

\end{document}